\documentclass{article}
\usepackage{arxiv}
\usepackage[
    breaklinks=true,
    letterpaper=true,
    colorlinks,
]{hyperref}
\usepackage{times}
\usepackage[OT1]{fontenc}
\usepackage{microtype}
\usepackage{epsfig}
\usepackage{amsmath,mathtools}
\usepackage{amssymb}
\usepackage{booktabs,adjustbox}
\usepackage{graphicx,placeins}
\usepackage{multirow}
\usepackage{xcolor}
\usepackage[capitalize]{cleveref}
\usepackage{subcaption}
\usepackage{algorithm}
\usepackage{algpseudocode}
\graphicspath{{figures/}}

\newcommand{\authornote}[1]{\textsuperscript{\rm {#1}}}
\newcommand{\email}[1]{\href{mailto:#1}{\texttt{#1}}}
\newcommand*\samethanks[1][\value{footnote}]{%
    \footnotemark[#1]\hspace{0.4em}}

\newcommand{\repourl}{Available at \url{https://github.com/lafeat/ueraser}.}

\makeatletter
\DeclareRobustCommand\onedot{\futurelet\@let@token\@onedot}
\def\@onedot{\ifx\@let@token.\else.\null\fi}

\newcommand{\etc}{\emph{etc\@\onedot}}
\newcommand{\etal}{\emph{et~al\@\onedot}}
\newcommand{\ie}{\emph{i.e\@\onedot}}

\newcommand{\wrt}{\emph{w.r.t\@\onedot}}

\newcommand{\MethodVerb}{{UEraser}}
\newcommand{\Method}{\emph{\MethodVerb{}}}
\newcommand{\MethodLite}{\emph{\MethodVerb{}-Lite}}
\newcommand{\MethodMax}{\emph{\MethodVerb{}-Max}}

\newcommand{\ordinal}[1]{{#1}\textsuperscript{th}}

\DeclarePairedDelimiter{\parens}{\lparen}{\rparen}

\DeclarePairedDelimiter{\bracks}{[}{]}
\DeclarePairedDelimiter{\braces}{\{}{\}}

\DeclarePairedDelimiter{\norm}{\lVert}{\rVert}

\newcommand{\inputset}{\mathcal{X}}
\newcommand{\outputset}{\mathcal{Y}}
\newcommand{\cleanset}{\mathcal{D}_\textrm{clean}}
\newcommand{\testset}{\mathcal{D}_\textrm{test}}
\newcommand{\sampledist}{\mathcal{S}}
\newcommand{\augdist}{\mathcal{A}}
\newcommand{\ueset}{\mathcal{D}_\textrm{ue}}
\newcommand{\bx}{{\boldsymbol{x}}}
\newcommand{\by}{{\boldsymbol{y}}}
\newcommand{\btheta}{{\boldsymbol{\theta}}}
\newcommand{\bdelta}{{\boldsymbol{\delta}}}
\newcommand{\transform}{\mathcal{T}}
\newcommand{\loss}{\mathcal{L}}
\newcommand{\expect}{\mathbb{E}}

\title{%
    Learning the Unlearnable:
    \mbox{Adversarial Augmentations}
    Suppress
    \mbox{Unlearnable Example Attacks}}
\author{%
    Tianrui Qin\thanks{%
        Equal contribution.
        Correspondence to Xitong Gao
        (\email{xt.gao@siat.ac.cn}).
    }\hspace{0.5em}\authornote{a,b},
    Xitong Gao\samethanks[1]\hspace{0.1em}\authornote{a},
    Juanjuan Zhao\hspace{0.1em}\authornote{a},
    Kejiang Ye\hspace{0.1em}\authornote{a},
    Cheng-Zhong Xu\hspace{0.1em}\authornote{c} \\
    \authornote{a}\,%
        Shenzhen Institute of Advanced Technology,
        Chinese Academy of Sciences, China. \\
    \authornote{b}\,%
        University of Chinese Academy of Sciences, China. \\
    \authornote{c}\,%
        University of Macau, Macau S.A.R., China.
}
\date{}

\begin{document}
    \maketitle
    \setcounter{footnote}{0}
    \begin{abstract}
    Unlearnable example attacks
    are data poisoning techniques
    that can be used to safeguard public data
    against unauthorized training of deep learning models.
    These methods add stealthy perturbations
    to the original image,
    thereby making it difficult
    for deep learning models
    to learn from these training data effectively.
    Current research suggests
    that adversarial training
    can, to a certain degree,
    mitigate the impact of unlearnable example attacks,
    while common data augmentation methods
    are not effective against such poisons.
    Adversarial training, however,
    demands considerable computational resources
    and can result in non-trivial accuracy loss.
    In this paper,
    we introduce the \Method{} method,
    which outperforms current defenses
    against different types of state-of-the-art
    unlearnable example attacks
    through a combination
    of effective data augmentation policies
    and loss-maximizing adversarial augmentations.
    In stark contrast
    to the current SOTA adversarial training methods,
    \Method{}
    uses adversarial augmentations,
    which extends beyond the confines
    of \( \ell_p \) perturbation budget
    assumed by current unlearning attacks and defenses.
    It also helps
    to improve the model's generalization ability,
    thus protecting against accuracy loss.
    \Method{} wipes out
    the unlearning effect
    with error-maximizing data augmentations,
    thus restoring trained model accuracies.
    Interestingly,
    \MethodLite{},
    a fast variant without adversarial augmentations,
    is also highly effective
    in preserving clean accuracies.
    On challenging unlearnable
    CIFAR-10, CIFAR-100, SVHN,
    and ImageNet-subset datasets
    produced with various attacks,
    it achieves results that are comparable
    to those obtained during clean training.
    We also demonstrate the efficacy
    of \Method{}
    against possible adaptive attacks.
    Our code is open source and available to the deep
    learning community\footnote{\repourl}.
\end{abstract}

    \section{Introduction}\label{sec:intro}

\begin{figure*}[ht]
    \centering
    \includegraphics[width=\linewidth]{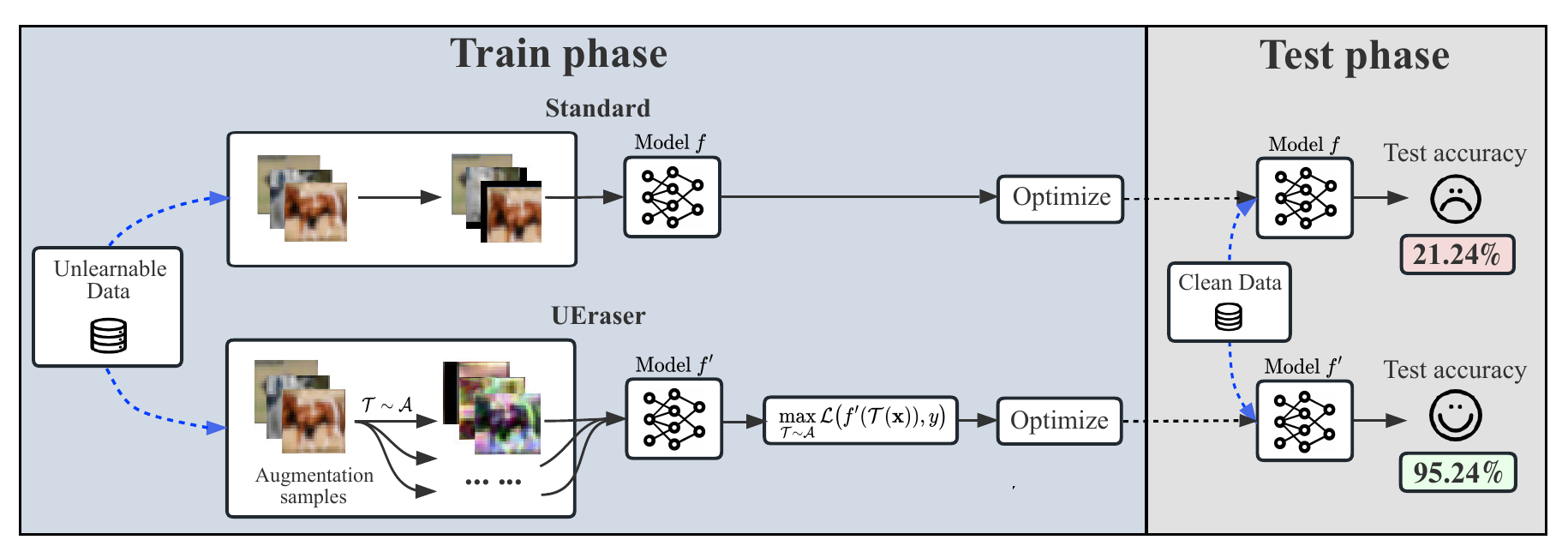}
    \caption{%
        A high-level overview of \Method{} for
        countering unlearning poisons.
        Note that \Method{} recovers the clean
        accuracy of unlearnable examples
        by data augmentations.
        The reported results
        are for EM~\cite{huangunlearnable}
        unlearnable CIFAR-10
        with an \( \ell_\infty \) perturbation budget
        of \( 8/255 \).
    }\label{fig:overview}
\end{figure*}
Deep learning has achieved great success in fields
such as computer vision~\cite{he2016deep}
and natural language processing~\cite{devlin2018bert},
and the development of various fields
now relies on large-scale datasets.
While these datasets
have undoubtedly contributed significantly
to the progress of deep learning,
the collection of unauthorized
private data for training these models
now presents an emerging concern.
Recently, numerous poisoning methods~\cite{
    furobust, huangunlearnable,
    sandoval2022autoregressive,
    tao2021better,
    yu2022availability}
have been proposed
to add imperceptible perturbations to images.
These perturbations
can form ``shortcuts''~\cite{%
    geirhos2020shortcut, huangunlearnable}
in the training data
to prevent training
and thus make the data unlearnable
in order to preserve privacy.
It is commonly perceived
that the only effective defense against unlearnable examples
are adversarial training algorithms~\cite{huangunlearnable,tao2021better,furobust}.
Popular data augmentation methods
such as CutOut~\cite{devries2017improved},
MixUp~\cite{zhang2017mixup},
and AutoAugment~\cite{cubuk2019autoaugment},
however,
have all been demonstrated to be ineffective defenses.


Current methods of unlearnable attacks
involves the specification
of an \( \ell_p \) perturbation budget,
where \( p \in \braces{2, \infty} \) in general.
Essentially,
they constrain the added perturbation
to a small \( \epsilon \)-ball of \( \ell_p \)-distance
from the source image,
in order to ensure stealthiness of these attacks.
Adversarial training defenses~\cite{madry2017towards,furobust}
represent a defense mechanism
that seeks to counteract the bounded perturbations
from such unlearnable attacks.
However, large defensive perturbations
comes with significant accuracy degradations.
This prompts the inquiry of the existence
of effective defense mechanisms
that leverage threat models
that are outside the purview of attackers.
Specifically,
\emph{can we devise effective adversarial policies
for training models
that extend beyond the confines
of the \( \ell_p \) perturbation budgets?}

In this paper,
we thus propose \Method{},
which performs error-maximizing data augmentation,
to defense against unlearning poisons.
\Method{} challenges
the preconception that data augmentation
is not an effective defense against unlearning poisons.
\Method{} expands the perturbation distance
far beyond traditional adversarial training,
as data augmentation policies
do not confine themselves
to the \( \ell_p \)
perturbation constraints.
It can therefore effectively disrupt ``unlearning shortcuts''
formed by attacks within narrow \( \ell_p \) constraints.
Yet, the augmentations employed by \Method{}
are natural and realistic transformations
extensively utilized by existing works
to improve the models' ability to generalize.
This, in turn, helps in avoiding accuracy loss
due to perturbations used by adversarial training
that could potentially be out-of-distribution.
Finally,
traditional adversarial training
is not effective in mitigating unlearning poisons
produced by adaptive attacks~\cite{furobust},
while \Method{} is highly resiliant
against adaptive attacks
with significantly lower accuracy reduction.

In summary, our work has three main contributions:
\begin{itemize}
    \item
    It extends adversarial training
    beyond the confines
    of the \( \ell_p \) perturbation budgets
    commonly imposed by attackers
    into data augmentation policies.

    \item
    We propose \Method{},
    which introduces an effective adversarial augmentation
    to wipe out unlearning perturbations.
    It defends against the unlearnable attacks
    by maximizing the error of the augmented samples.

    \item
    \Method{} is highly effective
    in wiping out the unlearning effect
    on five state-of-the-art (SOTA) unlearning attacks,
    outperforming existing SOTA defense methods.

    \item
    We explore the adaptive attacks
    on \Method{}
    and explored additional combinations
    of augmentation policies.
    It lays a fresh foundation for future competitions
    among unlearnable example attack and defense
    strategies.
\end{itemize}

Unlearnable example attacks bear great significance,
not just from the standpoint of privacy preservation,
but also as a form of data poisoning attack.
It is thus of great significance
to highlight the shortcomings
of current attack methods.
Perhaps most surprisingly,
even a well-known unlearnable attack
such as EM~\cite{huangunlearnable}
is unable to impede the effectiveness of \Method{}.
By training a ResNet-18 model from scratch
using exclusively CIFAR-10 unlearnable data
produced with EM
(with an \( \ell_\infty \) budget of \( 8/255 \)),
\Method{} achieves exceptional accuracy of \( 95.24\% \)
on the clean test set,
which closely matches the accuracy achievable
by standard training on a clean training set.
This suggests that existing unlearning perturbations
are tragically inadequate
in making data unlearnable,
even with adaptive attacks
that employs \Method{}.
By understanding their weaknesses,
we can anticipate how malicious actors
may attempt to exploit them,
and prepare stronger safeguards against such threats.
We hope \Method{}
can help facilitate the advancement
of research in these attacks and defenses.

    \section{Related Work}

\textbf{Adversarial examples and adversarial training.}
Adversarial examples deceive machine learning
models by adding adversarial perturbations,
often imperceptible to human,
to source images,
leading to incorrect classification
results~\cite{goodfellow2014explaining,szegedy2013intriguing}.
White-box adversarial attacks~\cite{szegedy2013intriguing}
maximize the loss of a source image
with gradient descent on the defending model
to add adversarial perturbations
onto an image to maximize its loss
on the model.
Effective methods to gain adversarial robustness
usually involve adversarial training~\cite{madry2017towards},
which leverages adversarial examples to train models.
Adversarial training algorithms
thus solve the min-max problem
of minimizing the loss function
for most adversarial examples
within a perturbation budget,
typically bounded in \( \ell_p \).
Recent years have thus observed an arms race
between adversarial attack strategies
and defense mechanisms~\cite{%
    croce2020robustbench,croce2020reliable,yu2021lafeat,yu2022mora}.

\textbf{Data poisoning.}
Data poisoning attacks
manipulate the training of a deep learning model
by injecting malicious and poisoned examples
into its training set~\cite{biggio2012poisoning,shafahi2018poison}.
Data poisoning methods~\cite{chen2017targeted,nguyen2021wanet}
achieve their malicious objectives
by stealthily replacing a portion
of training data,
and successful attacks can be triggered
with specially-crafted
prescribed inputs.
Effective data poisoning attacks
typically perform well on clean data
and fail on data that contains triggers~\cite{li2022backdoor}.

\textbf{Unlearnable examples.}
Unlearnable examples attacks
are a type of data poisoning methods
with bounded perturbation
that aims to make learning
from such examples difficult.
Unlike traditional data poisoning methods,
unlearnable examples methods usually
require adding imperceptible perturbations
to all examples~\cite{
    furobust,huangunlearnable,sandoval2022autoregressive,
    tao2021better,yu2022availability}.
Error-minimizing (EM)~\cite{huangunlearnable} poison
generates imperceptible perturbations
with a min-min objective,
which minimizes the errors of training examples
on a trained model,
making them difficult to learn by deep learning models.
By introducing noise
that minimizes the error of all training examples,
the model instead learns ``shortcut'' of such perturbations,
resulting in inability to learn from such data.
Hypocritical perturbations (HYPO)~\cite{tao2021better}
follows a similar idea
but uses a pretrained surrogate
rather than the above min-min optimization.
As the above method cannot defend
against adversarial training,
Robust Error-Minimizing (REM)~\cite{furobust}
uses an adversarially-trained model
as an adaptively attack
to generate stronger unlearnable examples.
INF~\cite{wen2023adversarial}
enables samples from different classes
to share non-discriminatory features
to improve resistance to adversarial training.
Linear-separable Synthetic Perturbations
(LSP)~\cite{yu2022availability}
reveals that if the perturbations
of unlearnable samples
are assigned to the corresponding target label,
they are linearly separable.
It thus proposes
linearly separable perturbations
in response to this characteristic
and show great effectiveness.
Autoregressive poisoning
(AR)~\cite{sandoval2022autoregressive}
proposes a generic perturbation
that can be applied to different
datasets and architectures.
The perturbations of AR
are generated from dataset-independent processes.

\textbf{Data augmentations.}
Data augmentation techniques
increase the diversity
of training data by applying
random transformations~\cite{krizhevsky2012alexnet}
(such as rotation, flipping, cropping, \etc{})
to images,
thereby improving the model's generalization ability.
Currently,
automatic search-based
augmentation techniques
such as TrivialAugment~\cite{muller2021trivialaugment}
and AutoAugment~\cite{cubuk2019autoaugment},
can further improve the performance of trained
DNNs by using a diverse set of augmentation policies.
TorMentor~\cite{nicolaou2022tormentor},
an image-augmentation framework,
proposes fractal-based data augmentation
to improve model generalization.
Current unlearnable example methods~\cite{
    furobust,huangunlearnable,sandoval2022autoregressive,
    tao2021better,yu2022availability}
demonstrate strong results
under an extensive range of data augmentation methods.
Despite prevailing beliefs
on their ineffectiveness against unlearnable examples,
\Method{} challenges this preconception,
as it searches for adversarial policies
with error-maximizing augmentations
and achieves the state-of-art defense performance
against existing unlearnable example attacks.

    \section{The \Method{} Defense}

\subsection{Preliminaries on Unlearnable Example Attacks and Defenses}

\textbf{Attacker.}
We assume the attacker has access
to the original data
they want to make unlearnable,
but cannot alter the training process~\cite{li2022backdoor}.
Typically,
the attacker attempts
to make the data unlearnable
by adding perturbations to the images
to prevent trainers
from using them to learn a
classifier that generalize well
to the original data distribution.
Formally,
suppose we have a dataset
consisting of original clean examples
\(
    \cleanset = \braces*{
        \parens*{\bx_{1}, y_{1}},
        \ldots,\parens*{\bx_{n}, y_{n}}
    }
\)
drawn from a distribution \( \sampledist \),
where \( \bx_i \in \inputset \) is an input image
and \( y_i \in \outputset \) is its label.
The attacker thus aims
to construct a set
of sample-specific unlearning perturbations
\( \bdelta = \{ \bdelta_\bx | \bx \in \inputset \} \),
in order to make the model
\( f_\btheta \colon \inputset \rightarrow \outputset \)
trained on the \emph{unlearnable examples} set \(
    \ueset\parens{\bdelta} = \braces*{
        \parens{\bx + \bdelta_\bx, y} \mid
        \parens{\bx, y} \in \cleanset
    }
\)
perform poorly on a test set \( \testset \)
sampled from \( \sampledist \):
\begin{equation}
    \max_{\bdelta}
    \expect_{
        (\bx_i, y_i) \sim \testset
    }\bracks{
        \loss\parens{
            f_{\btheta^\star\parens{\bdelta}} \parens{\bx_i},
            y_i
        }
    },
    \text{ s.t. }
    \btheta^\star\parens{\bdelta}
        = \underset{\btheta}{\operatorname{argmin}}
    \sum_{
        \qquad
        \mathclap{\parens*{\bx_{i}, y_{i}}
        \in \ueset \parens{\bdelta}}
        \qquad
    }
    \loss \parens*{
        f_{\btheta} \parens*{ \hat\bx_{i} },
        y_{i}
    },
    \label{eq:attacker}
\end{equation}
where \( \loss \) is the loss function,
typically the softmax cross-entropy loss.
For each image,
the noise \( \bdelta_i \) is bounded
by \( \norm{\bdelta_i}_p \leq \epsilon \),
where \( \epsilon \) is a small perturbation budget
such that it may not affect the intended utility of the image,
and \( \norm{\cdot}_p \) denotes the \( \ell_p \) norm.
\Cref{tab:visualization}
provides samples generated
by unlearnable example attacks
and their corresponding perturbations
(amplified with normalization).
\begin{table*}[!ht]
\centering
\caption{%
    The visualization of unlearned examples
    and perturbations of five poisoning methods on CIFAR-10.
}\label{tab:visualization}
\adjustbox{width=0.70\linewidth}{%
\begin{tabular}{c||ccc|cc}
\toprule
Clean
    & EM~\cite{huangunlearnable}
    & REM~\cite{furobust}
    & HYPO~\cite{tao2021better}
    & LSP~\cite{yu2022availability}
    & AR~\cite{sandoval2022autoregressive}\\
\midrule
\begin{minipage}[h]{0.15\columnwidth}
            \centering
            {\includegraphics[width=0.8\textwidth]{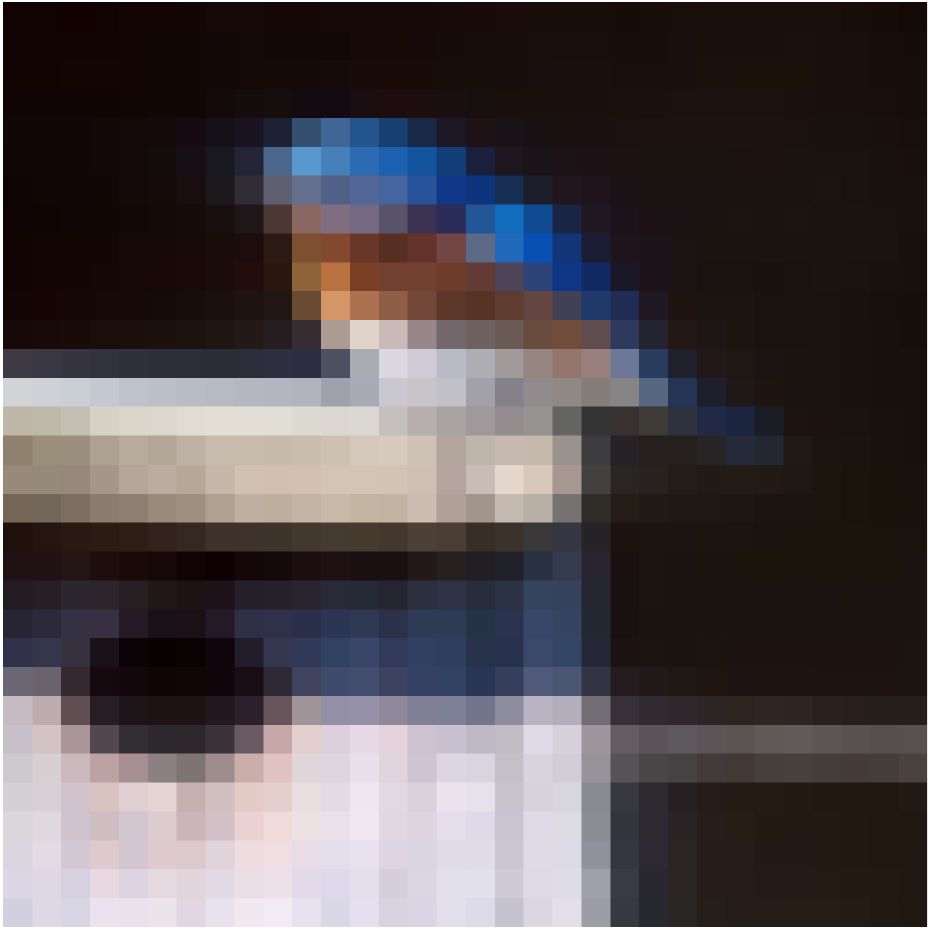}}
        \end{minipage}
        & \begin{minipage}[h]{0.15\columnwidth}
            \centering
            {\includegraphics[width=0.8\textwidth]{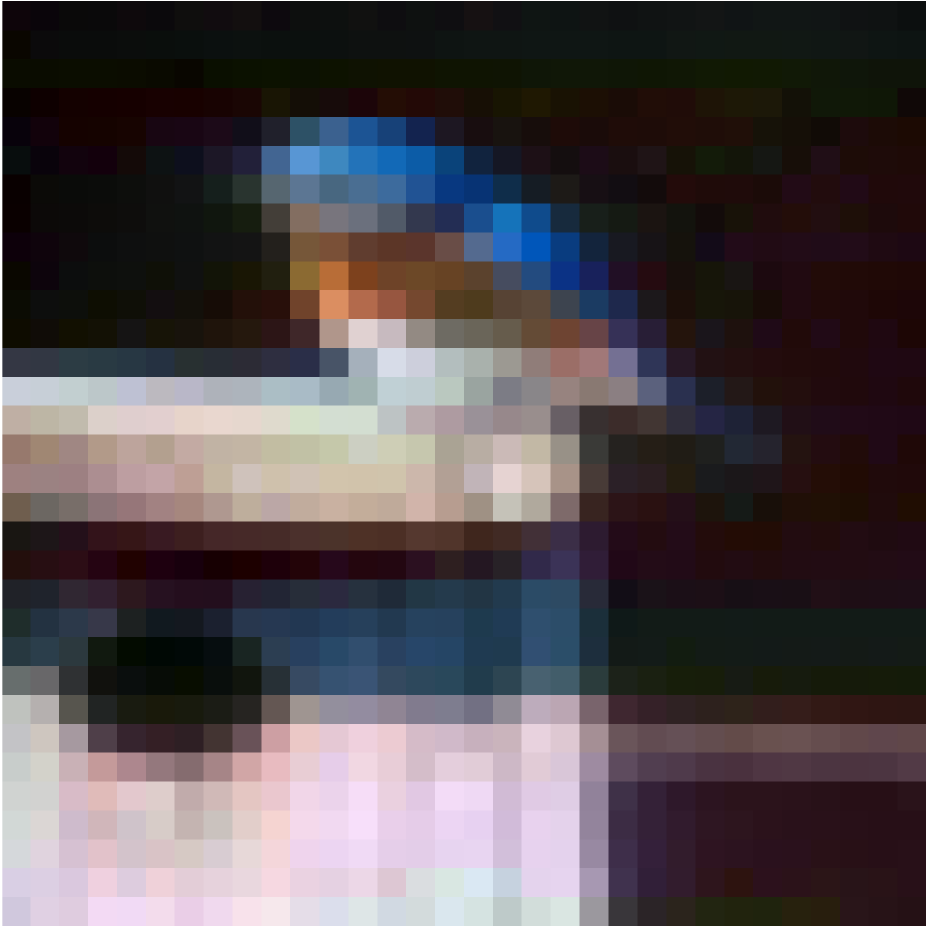}}
        \end{minipage}
        & \begin{minipage}[h]{0.15\columnwidth}
            \centering
            {\includegraphics[width=0.8\textwidth]{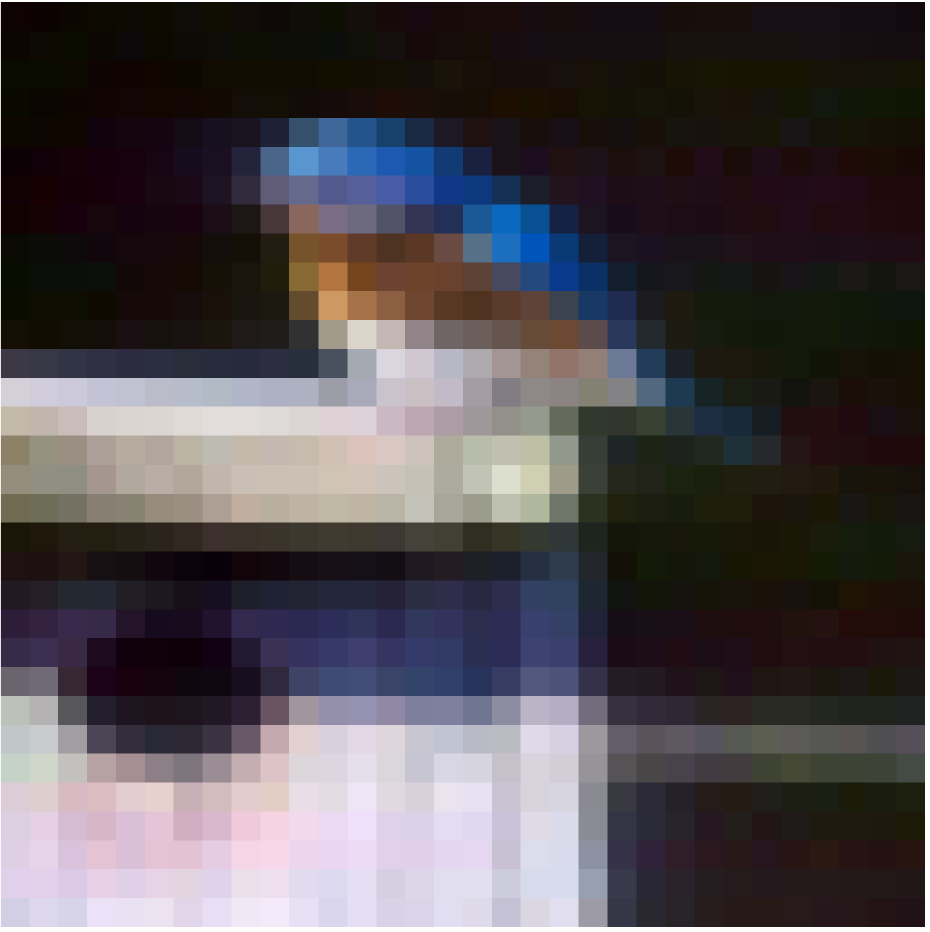}}
        \end{minipage}
        & \begin{minipage}[h]{0.15\columnwidth}
            \centering
            {\includegraphics[width=0.8\textwidth]{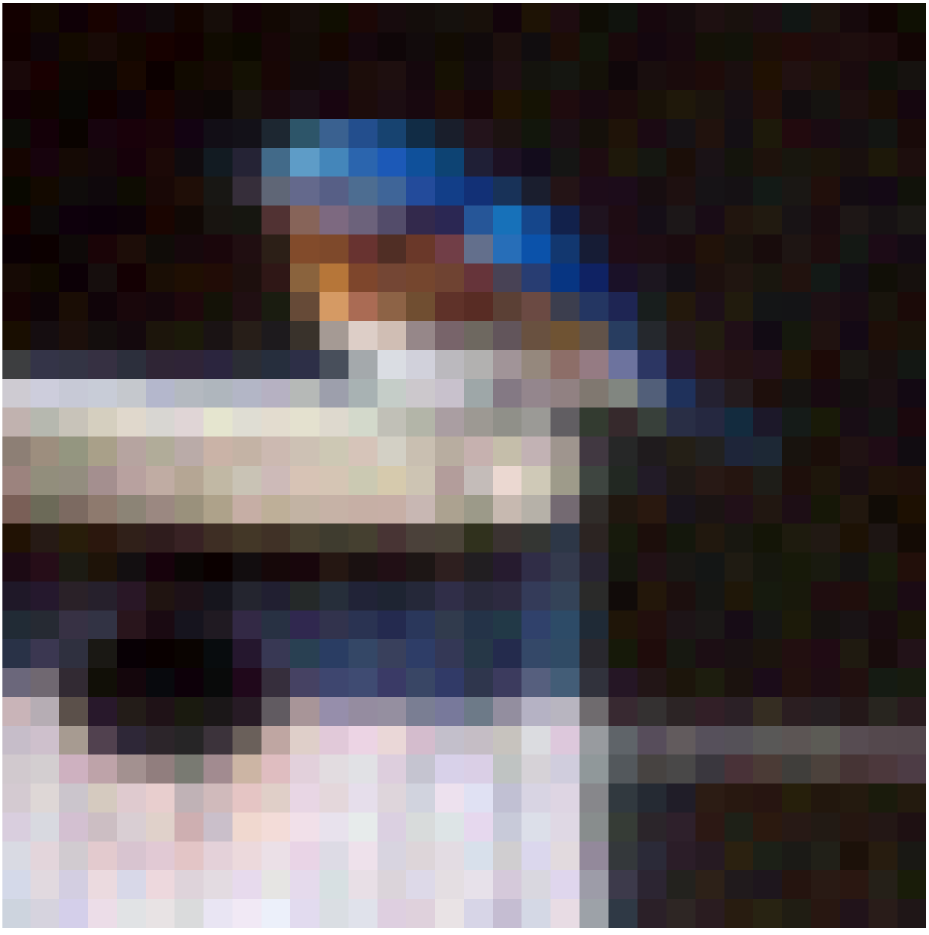}}
        \end{minipage}
        &\begin{minipage}[h]{0.15\columnwidth}
            \centering
            {\includegraphics[width=0.8\textwidth]{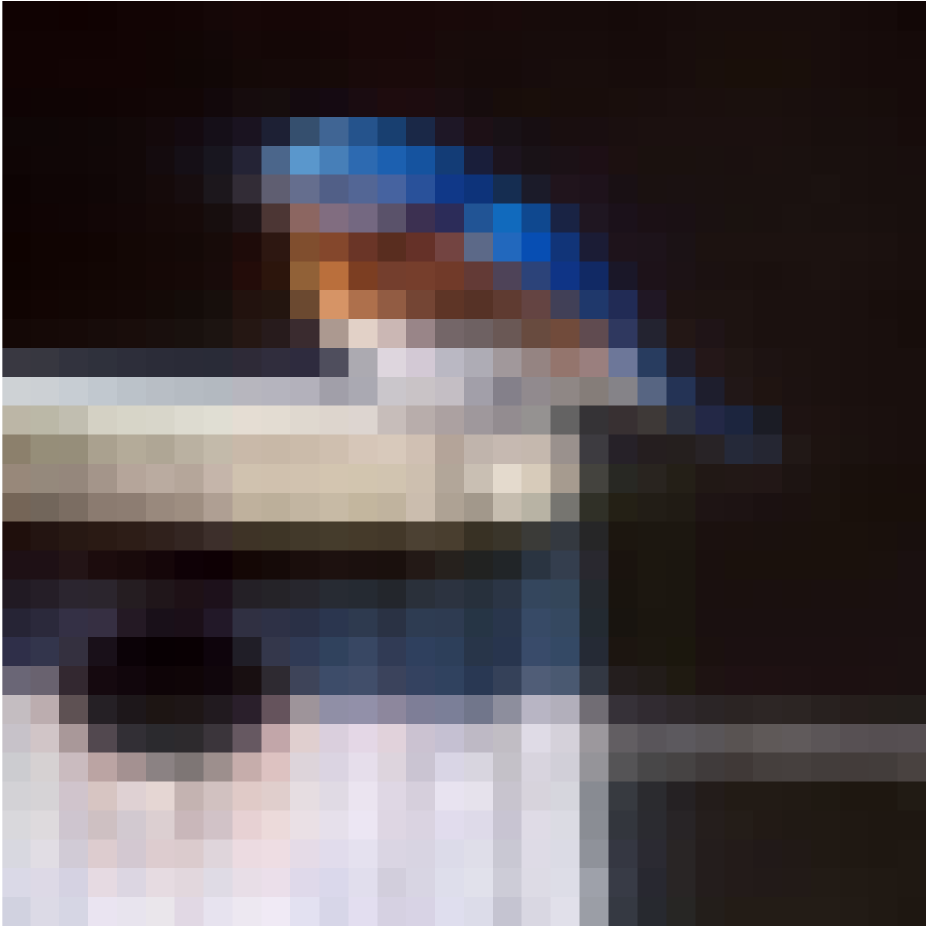}}
        \end{minipage}
        & \begin{minipage}[h]{0.15\columnwidth}
            \centering
            {\includegraphics[width=0.8\textwidth]{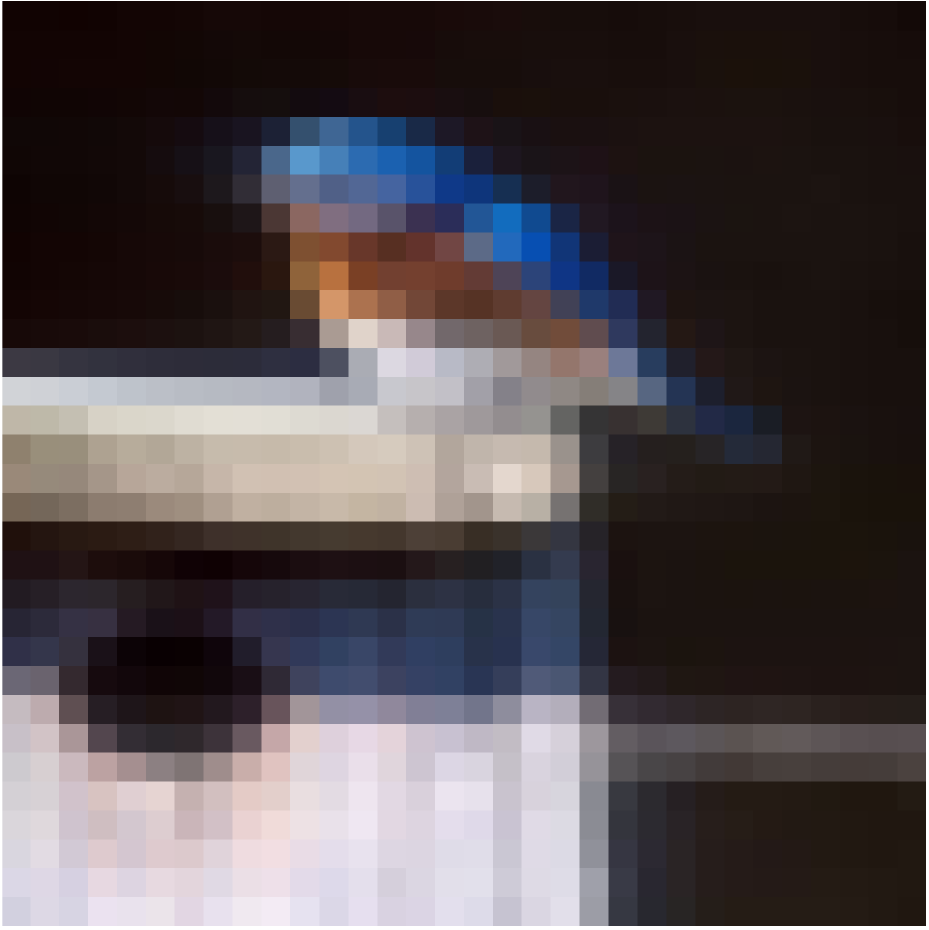}}
        \end{minipage} \\

\midrule
        Perturbations
        & \begin{minipage}[h]{0.15\columnwidth}
            \centering
            {\includegraphics[width=0.8\textwidth]{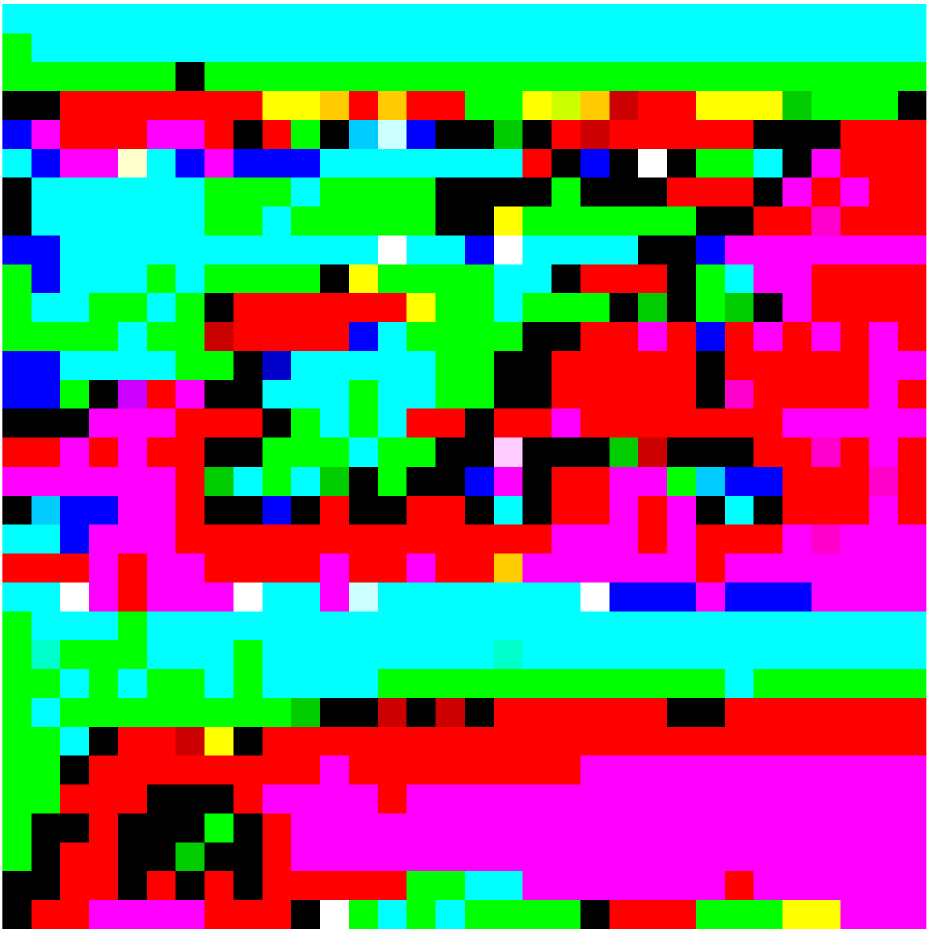}}
        \end{minipage}
        & \begin{minipage}[h]{0.15\columnwidth}
            \centering
            {\includegraphics[width=0.8\textwidth]{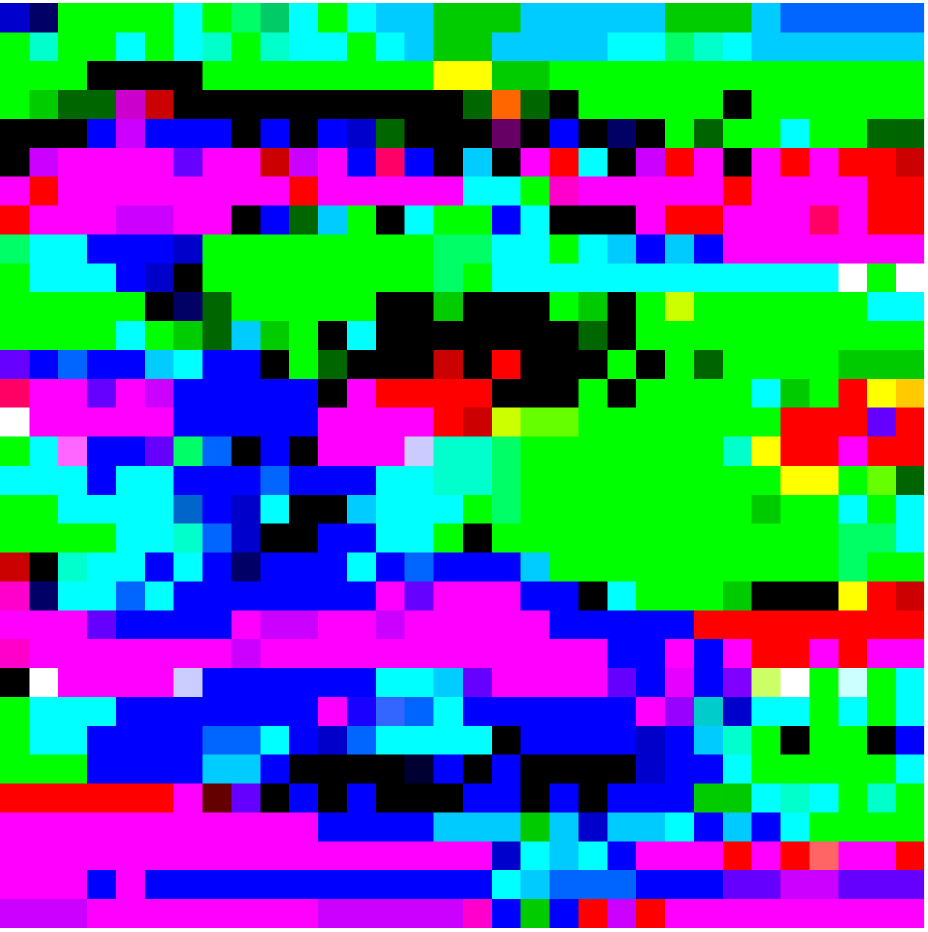}}
        \end{minipage}
        &\begin{minipage}[h]{0.15\columnwidth}
            \centering
            {\includegraphics[width=0.8\textwidth]{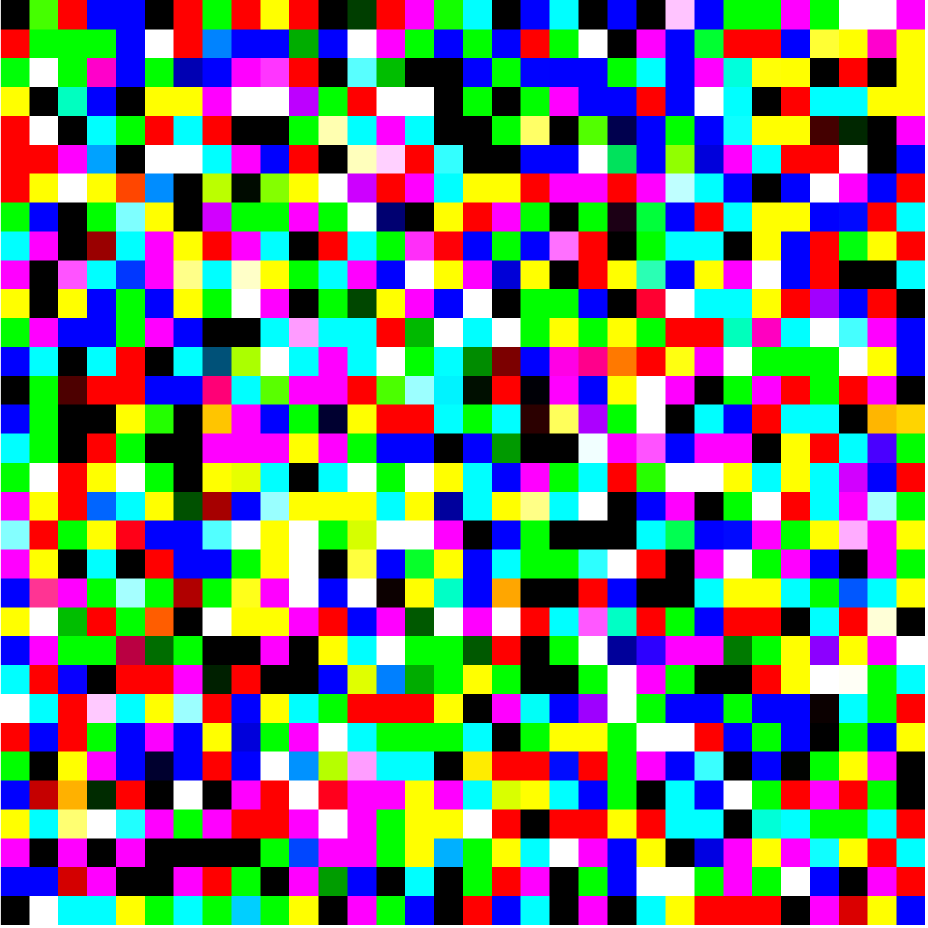}}
        \end{minipage}
        &\begin{minipage}[h]{0.15\columnwidth}
            \centering
            {\includegraphics[width=0.8\textwidth]{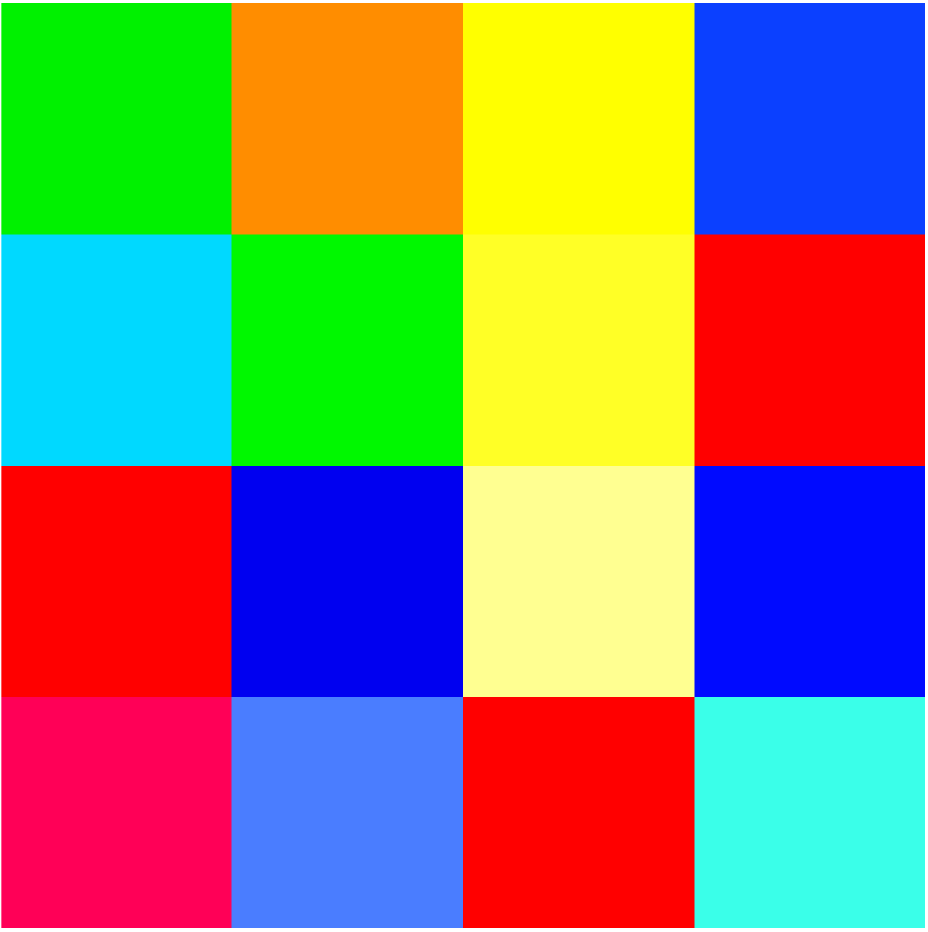}}
        \end{minipage}
        & \begin{minipage}[h]{0.15\columnwidth}
            \centering
            {\includegraphics[width=0.8\textwidth]{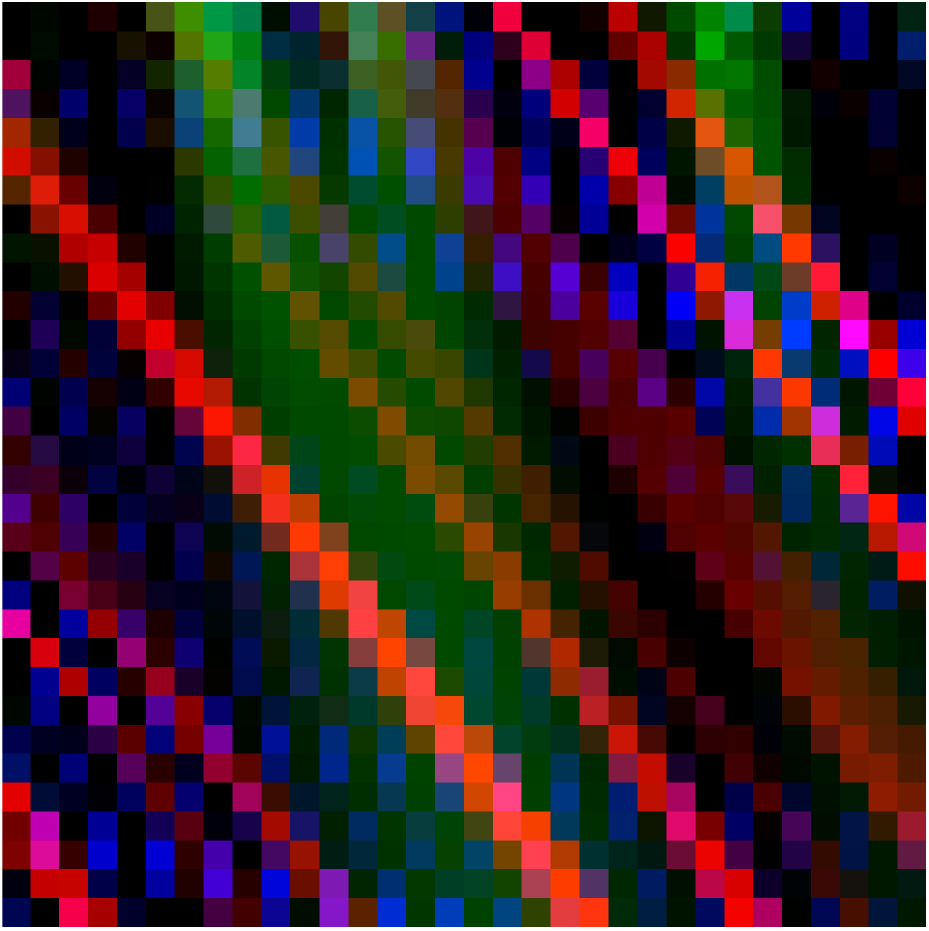}}
        \end{minipage} \\
\midrule
        Type
        & \multicolumn{3}{c|}{\( \ell_{\infty} \) bounded, \( \epsilon = 8/255 \)}
        & \multicolumn{2}{c}{\( \ell_{2} \) bounded, \( \epsilon = 1 \)}
          \\
\bottomrule
\end{tabular}}
\end{table*}

\textbf{Defender.}
The goal of the defender
is to ensure that the trained model
learns from the poisoned training data,
allowing the model to be generalized
to the original clean data distribution \( \cleanset \).
The attacker assumes full control
of its training process.
In our context,
we thus assume that the attacker's
policy is to perform poison removal on the image,
in order to ensure the trained model generalizes
even when trained on poisoned data \( \ueset \).
It has been shown
in~\cite{huangunlearnable,tao2021better,furobust}
that Adversarial training~\cite{madry2017towards}
is effective against unlearnable examples,
which optimizes the following objective:
\begin{equation}
    \arg\min_{\btheta}
    \expect_{\parens{\hat\bx, y} \sim \ueset} \bracks*{
        \max_{
            \norm{\bdelta_\textrm{adv}}_p \leq \epsilon
        }
        \loss\parens*{f_\btheta\parens{\hat\bx + \bdelta_\textrm{adv}}, y}
    }.
    \label{eq:adversarial_training}
\end{equation}
Specifically for each image \( \hat\bx \in \ueset \),
it finds an adversarial perturbation \( \bdelta_\textrm{adv} \)
that maximizes the classifier loss.
It then performs gradient descent on the maximal loss
to optimize for the model parameters \( \btheta \).
A model trained on the unlearnable set \( \ueset \) in this manner
thus demonstrates robustness
to perturbations in the input,
and can generalize to clean images.

\subsection{Adversarial Augmentations}

Geirhos~\etal{}~\cite{geirhos2020shortcut}
reveal that models tend to learn ``shortcuts'',
\ie{}, unintended features in the training images.
These shortcuts negatively
impact the model's generalization ability.
Intuitively,
unlearnable example attacks
thus leverage such shortcuts
to generate effective perturbations
to impede learning from the poisoned examples.
Subsequently,
existing adversarial training defenses~\cite{%
    huangunlearnable,tao2021better,furobust}
attempt to remove these shortcuts
from training images
with adversarial perturbations.
This is done to counter the effects
of the unlearning perturbations.

It is natural to think
that augmentation policies may be a dead end
against unlearnable attacks,
as none of the existing strong data augmentation methods
show significant effectiveness (\Cref{tab:ueaug}).
Adversarial training
can also be viewed as a practical
data augmentation policy,
which presents an interesting perspective
as it allows the model
to choose its own policy in the form
of \( \ell_p \)-bounded perturbations adaptively.
However, it poses a considerable challenge
due to its use of large defensive perturbations,
often resulting in reduced accuracy.
This begs the question
of whether new defense mechanisms
can leverage \emph{unseen} threat models
that unlearnable attacks
may be unable to account for.

Inspired by this,
we introduce \Method{},
which performs adversarial augmentations polices
that preserves to the semantic information
of the images
rather than adding \( \ell_p \)-bounded adversarial noise.
Our objective
is a bi-level optimization,
where the inner level
samples image transformation policies
\( \transform(\cdot) \)
from a set of all possible augmentations \( \augdist \),
in order to maximize the loss,
and the outer level performs model training
with adversarial polices:
\begin{equation}
    \arg\min_{\btheta} \expect_{
        (\bx, y) \sim \ueset
    } \bracks*{
        \max_{\transform \sim \augdist}
        \loss\parens*{
            f_{\btheta}\parens{
                \transform\parens*{\bx}
            }, y
        }
    }.
    \label{eq:method}
\end{equation}
Intuitively,
\Method{} finds the most ``adversarial''
augmentation policies
for the current images,
and use that to train the model
in order to prevent unlearnable ``shortcuts''
from emerging during model training.
Compared to adversarial training methods
that confine the defensive perturbations
within a small \( \epsilon \)-ball
of \( \ell_p \) distance,
here we adopt a different approach
that allows for a more aggressive distortion.
Moreover,
augmentation policies
also effectively preserve the original semantics
in the image.
By maximizing the adversarial loss in this manner,
the model can thus avoid training from the unlearning ``shortcuts''
and instead learn from the original features.

To generate augmentation policies
with high-level of distortions,
we select PlasmaTransform~\cite{nicolaou2022tormentor},
and TrivialAugment~\cite{muller2021trivialaugment},
two modern suites of data augmentation policies,
and ChannelShuffle in sequence,
to form a strong pipeline of data augmentations polices.
PlasmaTransform performs image distortion
with fractal-based transformations.
TrivialAugment
provide a suite of natural augmentations
which shows great generalization abilities
that can train models with SOTA accuracies.
Finally,
ChannelShuffle swaps the color channels randomly,
this is added to further increase the aggressiveness
of adversarial augmentation policies.
Interestingly,
using this pipeline
without the error-maximization augmentation sampling
can also significantly
reduce the effect of unlearning perturbations.
We denote this method as \MethodLite{},
as it requires only 1 augmentation sample
per training image.
Compared to \Method{},
although \MethodLite{}
may not perform as well as \Method{}
on most datasets,
it is more practical
than both \MethodLite{} and adversarial training
due to its faster training speed.

Finally,
we provide an algorithmic overview
of \Method{} in~\Cref{alg:method}.
It accepts as input
the poisoned training dataset \( \ueset \),
batch size \( B \),
randomly initialized model \( f_\btheta \),
number of training epochs \( N \),
number of error-maximizing augmentation epochs \( W \),
learning rate \( \alpha \),
number of repeated sampling \( K \),
and a suite of augmentation policies \( \augdist \).
For each sampled mini-batch \( \bx, \by \)
of data points from the dataset,
it applies \( K \) different
random augmentation policies
for each image in \( \bx \),
and compute the corresponding loss values
for all augmented images.
It then selects
for each image in \( \bx \),
the maximum loss produced by its \( K \) augmented variants.
The algorithm then uses the averaged loss
across the same mini-batch
to perform gradient descent
on the model parameters.
Finally,
the algorithm
returns the trained model parameters \( \btheta \)
after completing the training process.
\begin{algorithm}[h]
\caption{Training with \Method{}.}\label{alg:method}
\algnewcommand{\IfThen}[2]{
    \State{\algorithmicif\ {#1}\ \algorithmicthen\ {#2}}
    \algorithmicend\ \algorithmicif}
\newcommand{\algcmt}{\algorithmiccomment}
\begin{algorithmic}[1]
\Function{\MethodVerb{}}{%
    \( \ueset, B, f_\btheta, N, W, \alpha, K, \augdist \)~\!}
\For{\( n \in \bracks{1, \ldots, N} \)}
\IfThen{\( n \geq W \)}{\( K \gets 1 \)}
\algcmt{Disable adversarial augmentations after warmup.}
\For{\( (\bx, \by) \sim \mathrm{minibatch}(\ueset, B) \)}
\algcmt{Mini-batch sampling.}
\For{\( i \in \bracks{1, \ldots, B} \)}
\algcmt{For each image in mini-batch\ldots}
\For{\( j \in \bracks{1, \ldots, K} \)}
\algcmt{\ldots repeat \( K \) augmentations.}
    \State{\( \mathrm{aug} \sim \augdist \)}
    \algcmt{Sample augmentation policy\ldots}
    \State{\(
        \mathbf{L}_{ij} \gets \loss\parens{
            f_\btheta \parens{\mathrm{aug}(\bx_i)}, \by_i
        }
    \)}
    \algcmt{\ldots evaluate the loss function for the augmented image.}
\EndFor{}
\State{\(
    \mathbf{L}^\mathrm{adv}_{i}
    \gets \max_{j \in \bracks{1, \ldots, K}} \mathbf{L}_{ij}
\)}
\algcmt{Find the augmented image with maximum loss.}
\EndFor{}
\State{\(
    \btheta \gets \btheta -
        \alpha \nabla_{\btheta}
        \frac1B \sum_{i \in \bracks{1, \ldots, B}}
            \mathbf{L}^\mathrm{adv}_i
\)}
\algcmt{Gradient decent on the mini-batch of max-loss images.}
\EndFor{}
\EndFor{}
\State{\Return \(\btheta\)}
\EndFunction
\end{algorithmic}
\end{algorithm}

    \section{Experimental Setup \& Results}

\textbf{Datasets.}
We select four popular datasets for the
evaluation of \Method{}, namely,
CIFAR-10~\cite{krizhevsky2009learning},
CIFAR-100~\cite{krizhevsky2009learning},
SVHN~\cite{netzer2011reading},
and an ImageNet~\cite{deng2009imagenet} subset.
Following the setup in EM~\cite{huangunlearnable},
we use the first 100 classes of the
full dataset as the ImageNet-subset,
and resize all images to \( 32 \times 32 \).
We evaluate the effectiveness of \Method{}
by examining the accuracies of the trained models
on clean test examples,
\ie{}, the higher the clean test accuracy,
the greater its effectiveness.
By default,
all target and surrogate models
use ResNet-18~\cite{he2016deep}
if not otherwise specified.
We explore the effect of \Method{}
on partial poisoning (\Cref{sec:results:ablation:partial}),
larger perturbation budgets (\Cref{sec:results:ablation:larger_perturbation}),
different network architectures
(\Cref{sec:results:ablation:architectures}),
transfer learning (\Cref{sec:results:ablation:transfer}),
and perform ablation analyses
in~\Cref{sec:results:ablation:options,sec:results:ablation:repeat}.
Finally,
\Cref{app:sensitivity}
provides additional sensitivity analyses.

\textbf{Training settings.}
We train the CIFAR-10 model for 200 epochs,
SVHN model for 150 epochs,
and CIFAR-100 and ImageNet-subset models
for 300 epochs due to the use
of strong augmentation policies.
We adopt standard random cropping and flipping
for all experiments by default
as standard training baselines
and introduce additional augmentations
as required by the compared methods.
For the optimizer,
we use SGD with a momentum of 0.9
and a weight decay of \( 5 \times 10^{-4} \).
By default,
the learning rate is fixed at 0.01.
We follow the dataset setup in respective attacks,
where all unlearning perturbations
are bounded within \( \ell_\infty = 8/255 \)
or \( \ell_2 = 1.0 \).
For \Method{},
we divided the training process
into two parts for speed:
the adversarial augmentation process,
and the standard training process.
In the first stage,
we used the loss-maximizing augmentations for training,
with a default number of repeated samples
\( K = 5 \) (as the input to~\Cref{alg:method}).
In the second stage,
we used the \MethodLite{} process
which sets \( K = 1 \).
This approach allows us
to keep a balance
between suppressing the emergence of unlearning shortcuts
and training speed.
We further explore
full training with loss-maximizing augmentation
in~\Cref{sec:results:ablation:repeat},
which attains the highest known test accuracies.
Finally, \Cref{tab:main}
shows the effect of \Method{}
on five different unlearnable methods.
Additional information
regarding the training setup and hyperparameters
can be found in~\Cref{app:setup}.
The details of the attack and defense baselines
are available in~\Cref{app:baselines}.
\begin{table*}[!h]
\centering
\caption{%
    Clean test accuracies (\( \% \)) of \Method{} on CIFAR-10.
    All experiments are conducted on the ResNet-18 architecture.
    When training on clean data with \MethodLite{}
    and \Method{} the accuracy is 93.94\(\%\) and 93.66\(\%\) respectively.
    Note that ISS~\cite{liu2023image} contains three strategies
    (Grayscale, JPEG, and BDR),
    and we report the results of their best strategy.
    More specifically, Grayscale for EM and REM,
    JPEG for HYPO, LSP, and AR.
    For reference,
    we show the accuracies for ``Adversarial Training''
    with clean training samples in ``Clean''.
}\label{tab:main}
\begin{tabular}{c|cc|cc|c|cc}
\toprule
     \multirow{2}{*}{Methods}
    & \multicolumn{2}{c|}{Standard Training}
    & \multirow{2}{*}{\MethodLite{}}
    & \multirow{2}{*}{\Method{}}
    & \multirow{2}{*}{ISS~\cite{liu2023image}}
    & \multicolumn{2}{c}{Adversarial Training}\\

    & Clean & Unlearnable &  &  &  & Clean  & Unlearnable \\

\midrule

     EM
    & \multirow{5}{*}{94.78}
    & 21.24 & 90.78 &\textbf{93.38} &78.05 &88.71 &86.24\\

    REM&  &33.12 &85.49 &\textbf{91.02} &80.78 &87.15 &49.17\\

    HYPO&  &72.12 &85.67 &87.59 &84.77 &91.45 &\textbf{88.90}\\

    LSP&  &14.95 &84.92 &\textbf{85.07} &82.71 &81.24 &80.15\\

    AR&  &12.04 &92.08 &\textbf{93.16} &84.67 &81.09 &81.28\\
\bottomrule
\end{tabular}
\end{table*}

\subsection{Main Evaluation}

To demonstrate the effectiveness of \Method{},
we select five SOTA unlearnable example attacks:
Error-Minimization (EM)~\cite{huangunlearnable},
Hypocritical Perturbations (HYPO)~\cite{tao2021better},
Robust Error-Minimization (REM)~\cite{furobust},
Linear-separable Synthetic Perturbations (LSP)~\cite{yu2022availability},
and Autoregressive Poisoning (AR)~\cite{sandoval2022autoregressive}.
Experimental results show that \Method{}
has achieved better results
than the SOTA defense methods:
Image shortcut squeezing (ISS)~\cite{liu2023image}
and adversarial training~\cite{madry2017towards}.
For EM, REM, and HYPO,
We use the same model (ResNet-18)
as the surrogate and target model.
All unlearnable methods
have a poisoning rate of \( 100\% \).
From the experimental results of \Method{}
on CIFAR-10 dataset
shown in~\Cref{tab:main},
we can conclude
that \Method{} can achieve better defensive results
than ISS and adversarial training in most cases.
Note that the adversarial training experiments
use the same type of perturbation,
which is equal to half the size
of the unlearned perturbation.
Specifically,
the perturbation radius is \( \ell_\infty = 4/255 \)
for EM, REM, and HYPO,
and \( \ell_2 = 0.5 \)
for LSP and AR.
We consider sample-wise perturbations
for all experiments.
\MethodLite{} contains a series of augmentations
that effectively break through unlearned perturbation
while also slightly affecting the clean accuracy.
Therefore,
applying the \MethodLite{}
augmentation on clean data
may even lead to accuracy degradation,
and this degradation is further increased
when applying the \Method{}.  
When error-maximizing augmentation
is used throughout the training phase,
the model requires more epochs to converge
but achieves a higher accuracy rate (\(95.24\%\)).

In~\Cref{tab:extra},
We further validate the effect of \Method{}
on CIFAR-100, SVHN and ImageNet-subset.
We select the popular method (EM)
and the latest attacks for the experiments (LSP and AR).
Note that since \Method{}
increases the diversity of the data
with strong augmentations,
it requires more training epochs
to achieve converged accuracies.
All results are thus evaluated
after 300 training epochs
for CIFAR-100 and ImageNet-subset
and 150 training epochs for SVHN\@.
Experimental results
show that \Method{} achieves SOTA results
on all three datasets.
\begin{table}[!h]
\centering
\caption{
Clean test accuracies (\%) of ResNet-18 models
trained on CIFAR-10 unlearnable examples with
various attack and augmentation combinations.
``Standard'' denotes standard random crop and flip augmentations.
}\label{tab:ueaug}
\adjustbox{max width=\linewidth}{%
\begin{tabular}{c||ccccc}
\toprule
Methods
    & Standard
    & CutOut~\cite{devries2017improved}
    & MixUp~\cite{zhang2017mixup}
    & CutMix~\cite{devries2017improved}\\
\midrule
    EM~\cite{huangunlearnable}
    & 21.21 & 19.30 & 58.51 &22.40 \\
    REM~\cite{furobust}
    & 25.44 & 26.54 &29.02 &34.48  \\
    HYPO~\cite{tao2021better}
    & 23.69 & 27.14 & 35.44 & 29.33  \\
    LSP~\cite{yu2022availability}
    & 17.78 & 10.67 & 41.52 & 23.84 \\
    AR~\cite{sandoval2022autoregressive}
    & 11.75 & 11.90 & 11.40 & 11.23 \\
\bottomrule
\end{tabular}}
\end{table}

\begin{figure*}[ht]
\centering
\includegraphics[width=0.8\linewidth]{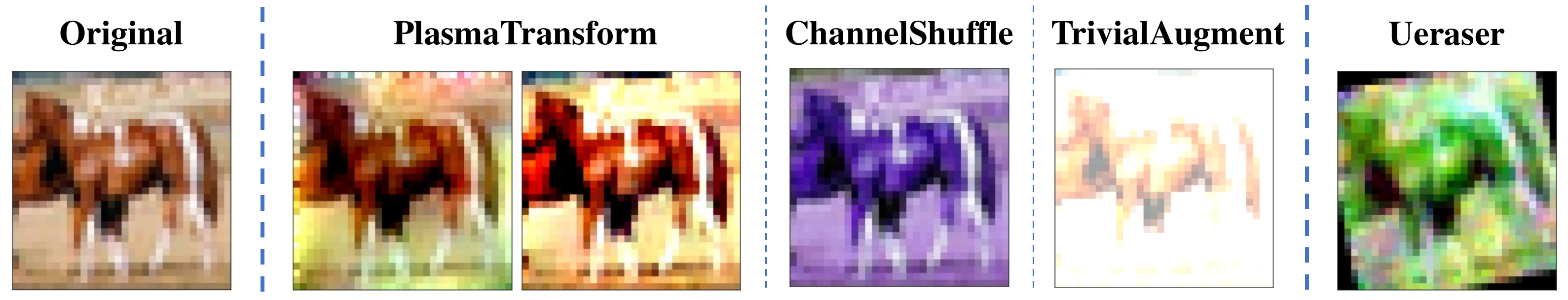}
\caption{%
    The Visualization of \Method{} augmentations.
}\label{fig:augvis}
\end{figure*}
\begin{table*}[!h]
\newcommand{\mbts}{\( {}^\dagger \)}
\centering
\caption{%
    Clean test accuracies (\( \% \)) of \Method{} on CIFAR-100,
    SVHN, and ImageNet-subset.
    The results of ISS~\cite{liu2023image}
    are from the best strategy
    (Grayscale for EM and JPEG for LSP).
    `\( \dagger \)' denotes the ImageNet-subset
    of the first 100 classes and resized to \(32 \times 32\).
}\label{tab:extra}
\begin{tabular}{c||cc|c|cccc}
\toprule
    {Dataset}
    & {Clean}
    & { +\MethodLite{}}
    & {Methods}
    & {Standard}
    & {\MethodLite{}}
    & {\Method{}}
    & {ISS~\cite{liu2023image}}
 \\

\midrule
    \multirow{3}{*}{CIFAR-100}
    & \multirow{3}{*}{74.83}
    & \multirow{3}{*}{73.22}
    & EM
    & 13.15
    & 70.04 & \textbf{71.14} &54.92 \\
    &
    &
    & LSP
    & 4.09
    & 67.38 & \textbf{68.42} &51.35 \\
    &
    &
    & AR
    & 4.79
    & 66.74 & \textbf{69.35} &57.34 \\
\midrule
    \multirow{3}{*}{SVHN}
    & \multirow{3}{*}{96.12}
    & \multirow{3}{*}{95.85}
    & EM
    & 10.58
    & 88.73    &\textbf{91.22} &89.91  \\
    &
    &
    & LSP
    & 14.56
    & 92.64 & \textbf{92.79} &92.17 \\
    &
    &
    & AR
    & 10.24
    & 93.04 & \textbf{93.83} &91.89  \\
\midrule
    \multirow{2}{*}{ImageNet{\mbts}}
    & \multirow{2}{*}{78.67}
    & \multirow{2}{*}{77.94}
    & EM
    & 17.83
    & 70.02 &\textbf{71.47}  &55.44 \\
    &
    &
    & LSP
    & 10.04
    & \textbf{65.12} & 63.87  & 52.27\\
\bottomrule
\end{tabular}
\end{table*}

    \subsection{Partial Poisoning}\label{sec:results:ablation:partial}

\begin{figure*}[htbp]
\centering
\begin{subfigure}{0.24\linewidth}
    \includegraphics[width=\linewidth]{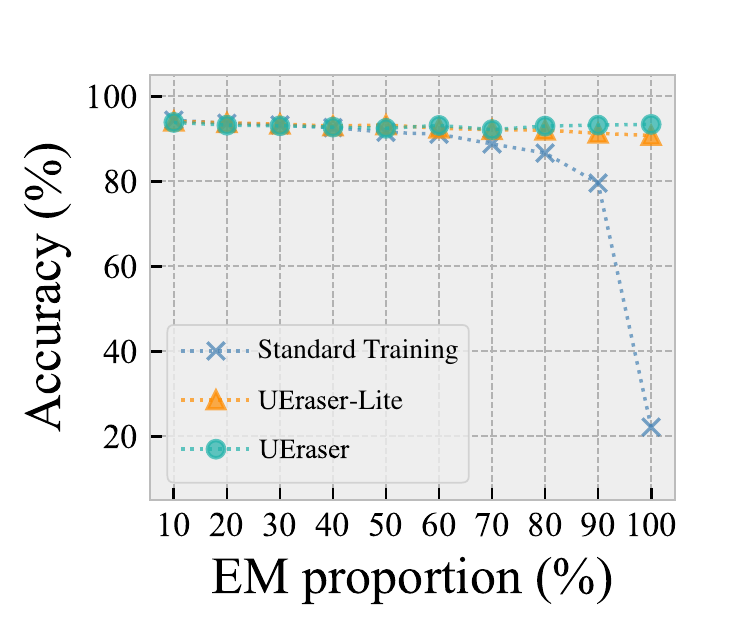}
    \caption{Partial poisoning of EM.}\label{fig:sensitivity:em}
\end{subfigure}
\hfill
\begin{subfigure}{0.24\linewidth}
    \includegraphics[width=\linewidth]{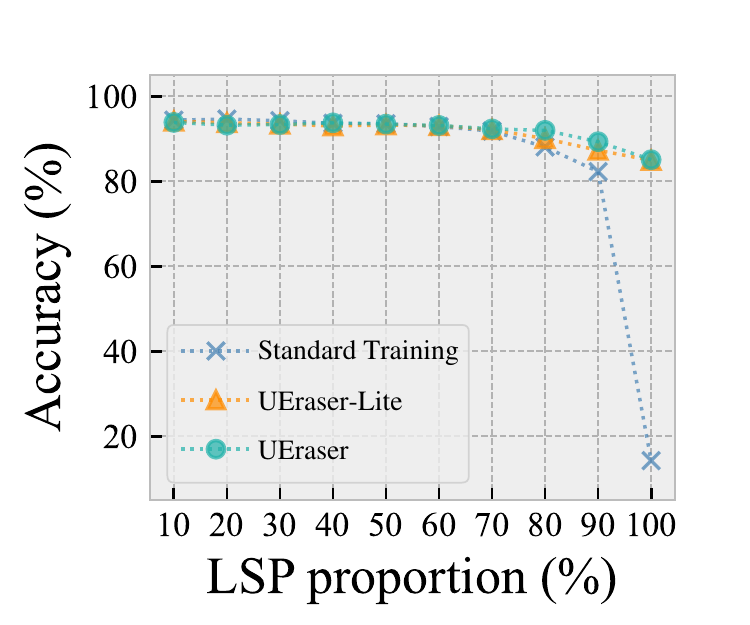}
    \caption{Partial poisoning of LSP.}\label{fig:sensitivity:lsp}
\end{subfigure}
\hfill
\begin{subfigure}{0.23\linewidth}
    \includegraphics[width=\linewidth]{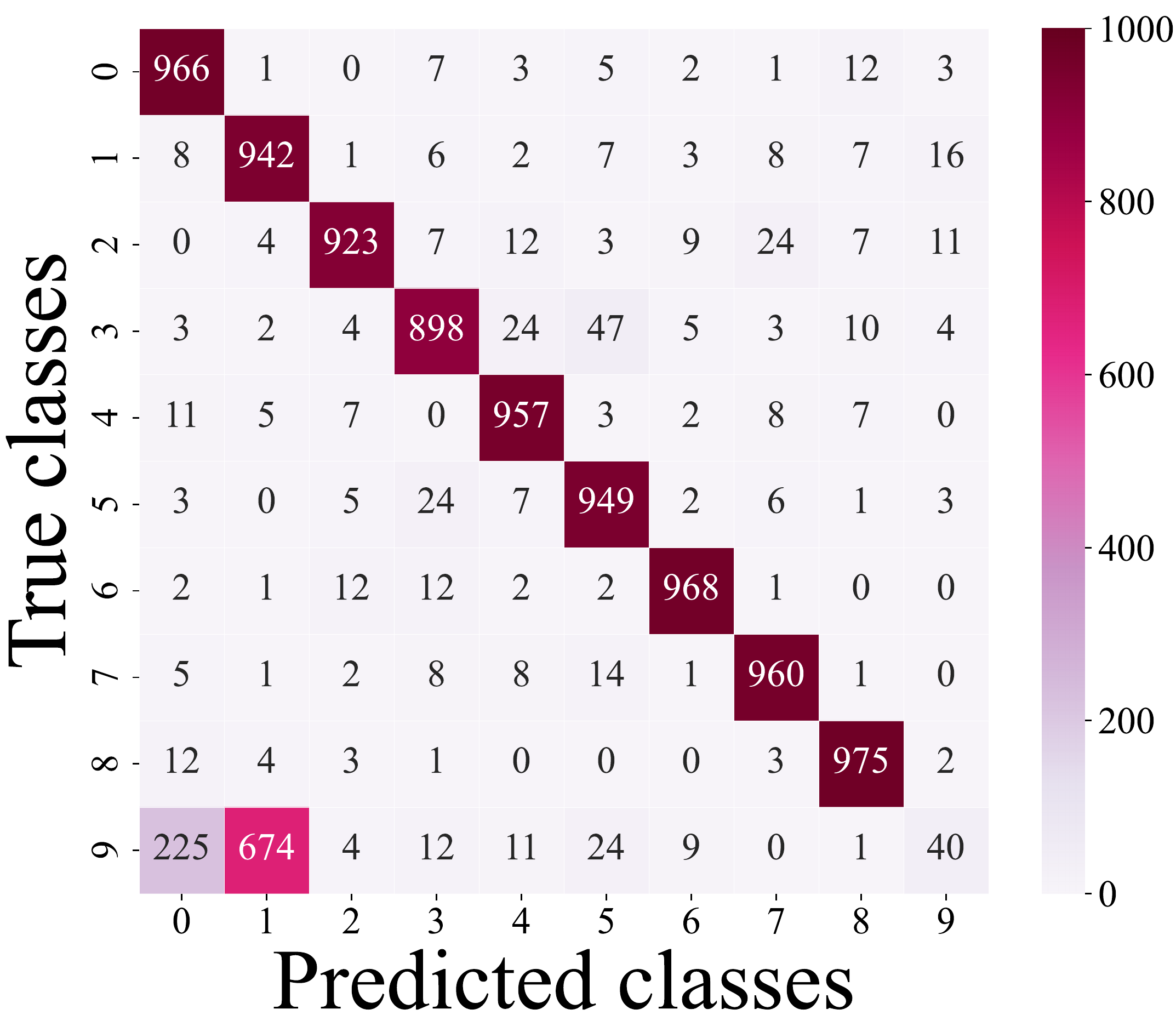}
    \caption{%
        Prediction matrix.
    }\label{fig:sensitivity:standard}
\end{subfigure}
\hfill
\begin{subfigure}{0.23\linewidth}
    \includegraphics[width=\linewidth]{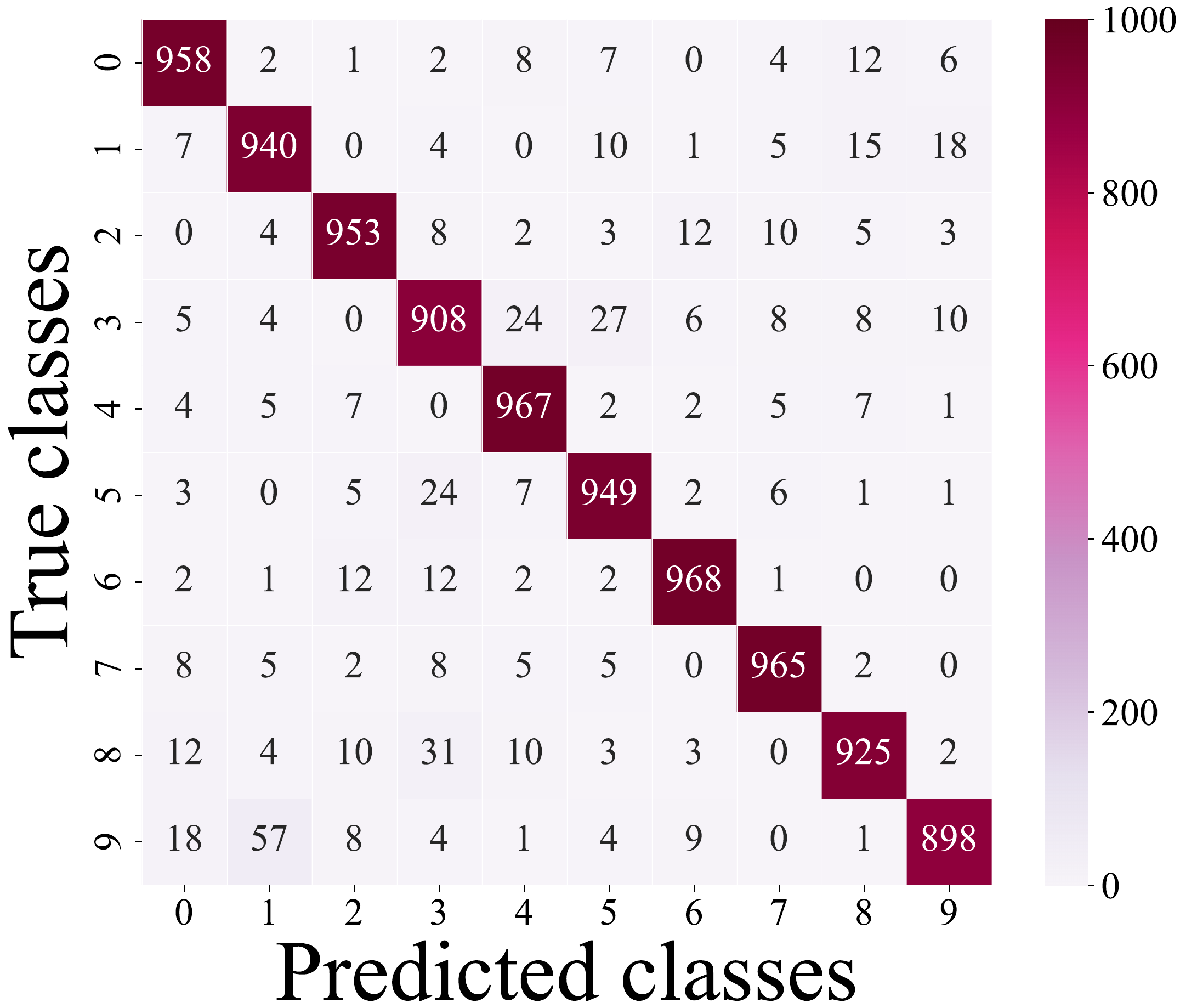}
    \caption{%
        With \Method{}.
    }\label{fig:sensitivity:ueraser}
\end{subfigure}
\caption{%
    The defensive efficacy of
    \Method{} against partial poisoning.
    (\subref{fig:sensitivity:em})
    EM with different poisoning ratios;
    (\subref{fig:sensitivity:lsp})
    LSP with different poisoning ratios.
    (\subref{fig:sensitivity:standard}),
    (\subref{fig:sensitivity:ueraser})
    Prediction confusion matrices
    on the clean test set
    of ResNet-18 trained on CIFAR-10
    with an unlearnable class
    (\emph{the \ordinal{9} label `truck'}).
    (\subref{fig:sensitivity:standard}) Standard training;
    (\subref{fig:sensitivity:ueraser}) \Method{} training.
}\label{fig:sensitivity}
\end{figure*}

In practical scenarios,
attackers may only have partial control
over the training data~\cite{huangunlearnable},
thus it is more practical
to consider the scenario
where only a part of the data is poisoned.
We adopt EM~\cite{huangunlearnable}
and LSP~\cite{yu2022availability}
on the CIFAR-10 dataset
as an example for our discussions.
Following the same setup,
we split varying percentages
from the clean data to carry out unlearnable poisoning
and mix it with the rest of the clean training data
for the target model training.
\Method{} is applied during model training
to explore its effectiveness
against partial poisoning.
\Cref{fig:sensitivity:em,fig:sensitivity:lsp},
show that when the poisoning ratio is low (\( <40\% \)),
the effect of the poisoning is negligible.
Another type of partial dataset attack scenario
is the selection of a targeted class to poison.
We thus poison all training
samples of the \ordinal{9} label (``truck''),
and \Cref{fig:sensitivity:standard,fig:sensitivity:ueraser}
shows the prediction confusion matrices
of ResNet-18 trained on CIFAR-10.
In summary,
\Method{} demonstrates significant efficacy
in partial poisoning scenarios.

\subsection{Adaptive Poisoning}\label{sec:results:ablation:adaptive}

Since \Method{} is composed
of multiple data augmentations,
we should consider possible adaptive
unlearnable example attacks
which may leverage \Method{}
to craft poisons against it.
We therefore evaluate \Method{}
in a worst-case scenario
where the adversary is fully aware
of our defense mechanism,
in order to reliably assess the resilience of \Method{}
against potential adaptive attacks.
Specifically,
we design an adaptive unlearning poisoning attack
by introducing an additional data augmentation
during the training,
We adopt the error-minimization
(EM) attack~\cite{huangunlearnable}
as an example.
The EM unlearning objective
solves the following min-min optimization:
\begin{equation}
    \arg\min_{\bdelta} \min_{\btheta}
    \expect_{(\bx, y) \sim \cleanset}\bracks*{
        \loss\parens*{
            f_\btheta \parens{\bx+\bdelta_{\bx}}, y
        }
    },
    \label{eq:attack:em}
\end{equation}
where \( \|\bdelta\|_{p} \leq \epsilon \).
Similar to the REM~\cite{furobust}
that generates adaptive unlearnable examples
under adversarial training,
each image \( \bx \)
optimizes its unlearning perturbation \( \bdelta_\bx \)
before performing adversarial augmentations:
\begin{equation}
    \arg\min_{\bdelta} \min_{\btheta}
    \expect_{(\bx, y) \sim \cleanset} \bracks*{
         \loss\parens*{
            f_\btheta \parens*{
                \transform_{\textrm{adv}}\parens{\bx + \bdelta_\bx}
            }, y
        }
    },
    \label{eq:attack:adaptive}
\end{equation}
where \( \transform_{\textrm{adv}}\parens{\cdot} \)
denotes the adversarial augmentations with \Method{}.
We select different combinations of augmentations
for our experiments,
and the results are shown in~\Cref{tab:adaptive}.
The hyperparameters of the augmentations
employed in all experiments
are kept consistent
with those of \Method{}.
We observe
that the adaptive augmentation of unlearning poisons
do not significantly reduce the effectiveness
of \Method{}.
As it encompasses a diverse set of augmentation policies
and augmentation intensities,
along with loss-maximizing augmentation sampling,
adaptive poisons are hardly effective.
Moreover,
we speculate that the affine and cropping transformations
in TrivialAugment can cause unlearning perturbations
to be confined to a portion of the images,
which also limits the effectiveness
of unlearning poisons.
Because of the aggressiveness of the augmentation
in image transformations
extend beyond the \( \ell_p \) bounds,
adaptive poisons do not perform as well under \Method{}
as they do against REM\@.
To summarize,
it is challenging for the attacker
to achieve successful poisoning
against \Method{}
even if it observes the possible transformations
taken by the augmentations.
\begin{table}[!h]
\centering
\caption{%
    Adaptive poisoning with EM
    on CIFAR-10.
    `P', `C,' and `T' denote
    PlasmaTransform, ChannelShuffle,
    and TrivialAugment respectively.
    ``Standard'' represents standard training.
}\label{tab:adaptive}
\begin{tabular}{l||ccc}
\toprule
    Methods & Standard & \MethodLite{} & \Method{} \\
\midrule
    Baseline & 21.21 & 90.78 & 93.38 \\
    \: + P & 24.36 & 86.05 &92.55   \\
    \: + C & 19.71 & 83.46 & 91.72   \\
    \: + P + C & 25.48 & 82.49 & 89.07 \\
    \: + P + C + T & 44.26 & 85.22 & 90.79 \\
\bottomrule
\end{tabular}
\end{table}

\subsection{%
    Larger Perturbation Scales
}\label{sec:results:ablation:larger_perturbation}

Will the performance of \Method{}
affected by large unlearnable perturbations?
To verify,
we evaluate the performance of \Method{}
on unlearnable CIFAR-10 dataset
with even larger perturbations.
We use the example
of error-maximizing (EM) attack
and increase the \( \ell_{\infty}\) perturbation
from \(8/255\) to \(24/255\)
to examine the efficacy of \Method{}
on a more severe unlearning perturbation scenario.
We also include adversarial training (AT)
as a defense baseline
with a perturbation bound of \( 8/255 \).
The experimental results in~\Cref{tab:scale}
confirm the effectiveness of \Method{}
under large unlearning noises.
\begin{table}[!h]
\centering
\caption{%
    Increasing the perturbation budget \( \epsilon \)
    of the EM attack
    on unlearnable defenses.
}\label{tab:scale}
\small
\begin{tabular}{r||cccc}
\toprule
Perturbation scale
    & Standard Training
    & \MethodLite{} & \Method{}
    & Adversarial Training \\
\midrule
    \( 8 / 255 \)  & 21.24 & 90.78 & \textbf{93.38} & 86.24 \\
    \( 16 / 255 \) & 22.63 & 86.65 & \textbf{89.24} & 83.12 \\
    \( 24 / 255 \) & 21.05 & 82.40 & \textbf{84.59} & 79.31 \\
\bottomrule
\end{tabular}
\end{table}

\subsection{%
    Resilience against Architecture Choices
}\label{sec:results:ablation:architectures}

Can \Method{} show resilience
against architecture choices?
In a realistic scenario,
we need to train the data
with different network architectures.
We thus explore whether \Method{}
can wipe out the unlearning effect
under different architectures.
\Cref{tab:arch} shows the corresponding results.
It is clear that \Method{}
is capable of effectively wiping out unlearning poisons,
across various network architectures.
\begin{table}[!h]
\centering
\caption{%
    Clean test accuracies (\%) of different architectures
    (ResNet-50~\cite{he2016deep},
     DenseNet-121~\cite{huang2017densely},
     MobileNet-V2~\cite{sandler2018mobilenetv2})
    on CIFAR-10.
    Note that EM is tested with \( \ell_{\infty} = 8 / 255 \)
    and LSP is tested with \( \ell_{2}=1\).
}\label{tab:arch}
\begin{tabular}{c|c||ccc}
\toprule
    \multicolumn{2}{c||}{Methods}
    & ResNet-50
    & DenseNet-121
    & MobileNet-V2
    \\
    \midrule
    \multicolumn{2}{c||}{Clean}
    & 94.39 & 95.14 & 94.20 \\

    \midrule
    \multirow{3}{*}{EM}
    & Standard
    & 25.17 & 34.91 & 31.75 \\
    & \MethodLite{}
    & 89.56 & 91.20 & 90.55 \\
    & \Method{}
    & 89.74 & 92.37 & 89.92 \\

    \midrule
    \multirow{3}{*}{LSP}
    & Standard
    & 14.94 & 22.71 & 20.04 \\
    & \MethodLite{}
    & 84.17 & 86.22 & 83.24 \\
    & \Method{}
    & 85.56 & 87.19 & 84.85 \\
\bottomrule
\end{tabular}
\end{table}

\subsection{Augmentation Options}\label{sec:results:ablation:options}

In this section,
we investigate the impact
of the \Method{} policy composition
on the mitigation of unlearning effects.
The visualization of the three policies
included is shown in~\Cref{fig:augvis}.
We conduct experiments
with the unlearnable examples from CIFAR-10
generated by the EM~\cite{huangunlearnable} method,
and~\Cref{tab:policy}
explore effectiveness of the various combinations
of augmentation policies.
ISS~\cite{liu2023image}
discovered that for the unlearnable examples
generated by the EM attack,
a grayscale filter easily
removes the unlearning poisons.
Additionally,
setting the value of each channel
to the average of all color channels
or to the value of any color channel
also considerately achieves the same effect.
However,
we show that using only ChannelShuffle
does not yield satisfactory results.
We have also discovered
an interesting phenomenon:
PlasmaTransform and ChannelShuffle
are essential for mostly restoring the accuracies,
whereas TrivialAugment,
can be substituted
with a similar policy,
\ie{}~AutoAugment~\cite{cubuk2019autoaugment}.
Only when the three policies
are employed together
can the effect of unlearning poisons
be effectively wiped out.
This also proves that the \Method{} policies
are effective and reasonable.
Recall that we also found the adoption
of error-maximizing augmentation
results in an overall improvement
on all five unlearning poisons.
Hence,
the utilization of error-maximizing
augmentation during training
serves as an effective means
to mitigate the challenges of training with unlearning examples
and improve the model's clean accuracy.
\begin{table}[t]
\centering
\caption{%
    Ablation analysis \Method{}
    on CIFAR-10.
    Note that all hyperparameters
    are the same,
    except for the \Method{}
    augmentation policy
    which varies.
    `P', `C', `T', and `A'
    denote ``PlasmaBrightness'', ``ChannelShuffle'',
    ``TrivialAugment'',
    and ``AutoAugment''~\cite{cubuk2019autoaugment}
    respectively.
    `Adv' denotes the full \Method{} method.
}
\begin{tabular}{lcc}
\toprule
Ablation of \Method{} Policies
    & Clean
    & Unlearnable (EM) \\
\midrule
Standard Training
    & 94.78 & 21.24 \\
\midrule
+ P
    & 94.47 & 29.48 \\
+ T
    & 94.47 & 48.81 \\
+ P + C
    & 94.15 & 62.17 \\
+ P + T
    & 95.22 & 48.05 \\
+ C + T
    & 94.40 & 69.24 \\
+ C + P + A
    & 94.04 & 85.60 \\
+ C + P + T
    & 93.94 & 90.78 \\
+ C + P + T + Adv
    & 93.66 & 93.38 \\

\bottomrule
\end{tabular}
\label{tab:policy}
\end{table}

\subsection{%
    Adversarial Augmentations and Error-Maximizing Epochs
}\label{sec:results:ablation:repeat}

From~\Cref{alg:method},
the training of \Method{}
is affected by two hyperparameters, namely,
the numbers of repeated augmentation samples \( K \)
per image
and the epochs of error-maximizing augmentations \( W \).
For CIFAR-10,
The clean accuracy of the unlearnable examples
can be improved to around \( 80\% \)
after 50 epochs of training
using error-maximizing augmentation.
We explored \MethodMax{}
which applies error-maximizing augmentations
throughout the entire training phase,
and it attains best known accuracies.
The results are shown in~\Cref{tab:ablation:emaug}.
Alternatively,
one can continue the training with \MethodLite{}
can improve the clean accuracy to \( \geq 93\% \)
to save computational costs.
Although \MethodMax{}
achieves the highest clean accuracy,
we mainly focus on \Method{} in this paper,
due to its high computational cost per epoch
and longer training epochs.
\begin{table}[t]
\centering
\caption{%
    Clean test accuracies (\%)
    of different \Method{} variants.
    Note that `200' and `300' denotes
    training for \( E = W = 200 \) and \( 300 \) epochs respectively.
}\label{tab:ablation:emaug}
\begin{tabular}{c|c|cccc}
\toprule
    \multirow{2}{*}{Methods}
    & \multirow{2}{*}{Clean}
    & \multirow{2}{*}{\MethodLite{}}
    & \multirow{2}{*}{\Method{}}
    & \multicolumn{2}{c}{\MethodMax{}}
    \\
    & & & &200 &300\\
    \midrule

    {EM}
    & \multirow{2}{*}{94.78}
    & 90.78 & 93.38 & 92.12 & \textbf{95.24} \\
    {LSP}
    &
    & 84.92 & 85.07 & 91.79 & \textbf{94.95} \\
\bottomrule
\end{tabular}
\end{table}

Regarding the number of samples \( K \)
(by default \( K = 5 \)),
increasing it
further enhances the suppression
of unlearning shortcuts
during model training,
but also more likely to lead to gradient explosions
at the beginning of model training.
Therefore,
it may be necessary
to apply learning rate warmup
or gradient clipping
with increased number of repeated sampling.
Larger \( K \) can also results
in higher computational costs,
as it result in more samples per image
for training.
We provide a sensitivity analysis
of the number of repeated sampling \( K \)
in~\Cref{app:sensitivity}.

\subsection{Transfer Learning}\label{sec:results:ablation:transfer}

In this section,
we aim to explore the impact
of transfer learning~\cite{dai2009eigentransfer,torrey2010transfer}
on the efficacy of unlearnable examples.
We hypothesize that pretrained models
may learn certain in-distribution features
of the unaltered target distribution,
it may be able to gain accuracy
even if the training set contains unlearning poisons.

To this end,
we adopt a simple transfer learning setup,
where we use the pretrained ResNet-18
model available in the torchvision repository~\cite{torchvision}.
To fit the expected input shape
of the feature extractor,
we upsampled the input images
to \( 224 \times 224 \).
The final fully-connected classification layer
of the pretrained model
was replaced with a randomly initialized one
with 10 logit outputs.
We then fine-tune the model with
unlearnable CIFAR-10 training data.
We also further explored fine-tuning
on unlearnable data with our defenses.
For control references,
we fine-tuned a model
with clean training data,
and also trained a randomly initialized model from scratch
with poisoned training data.
\begin{figure}[ht]
    \centering
    \begin{subfigure}{0.24\linewidth}
        \includegraphics[width=\linewidth]{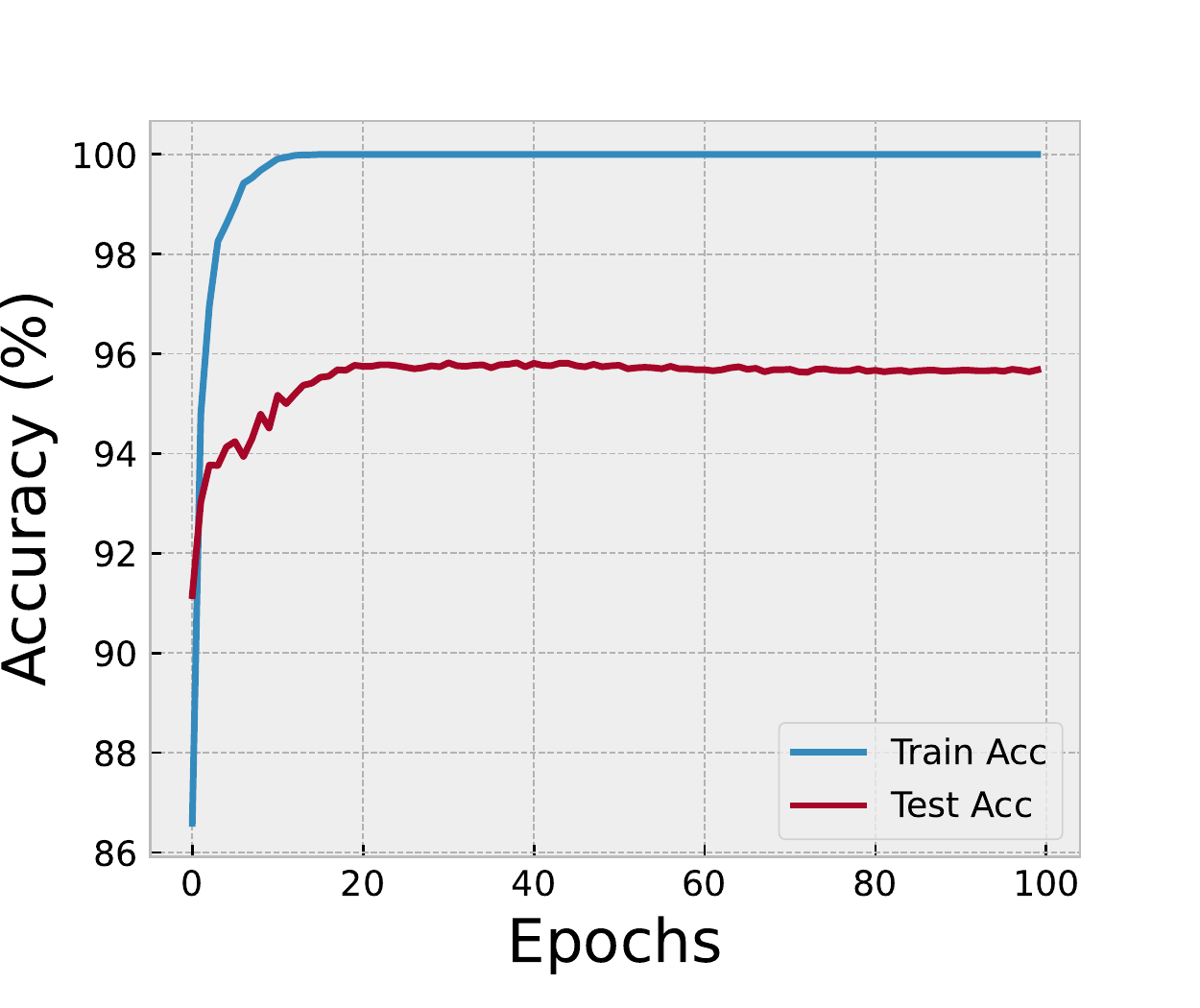}%
        \caption{%
            Fine-tuning on a clean training set.
        }\label{fig:transfer:clean}
    \end{subfigure}
    \hfill
    \begin{subfigure}{0.24\linewidth}
        \includegraphics[width=\linewidth]{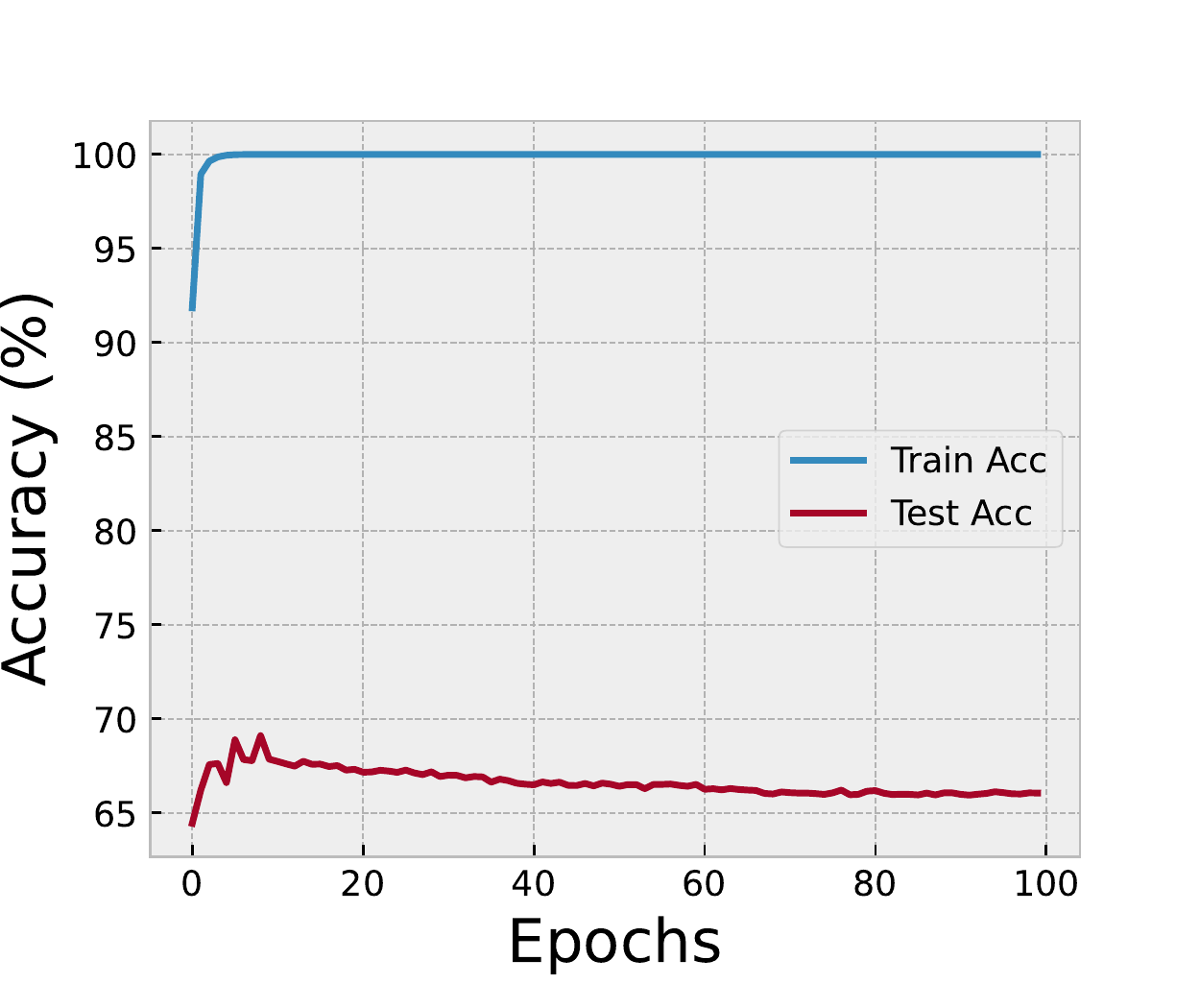}%
        \caption{%
            Fine-tuning
            on an unlearnable training set.
        }\label{fig:transfer:unlearn}
    \end{subfigure}
    \hfill
    \begin{subfigure}{0.24\linewidth}
        \includegraphics[width=0.93\linewidth]{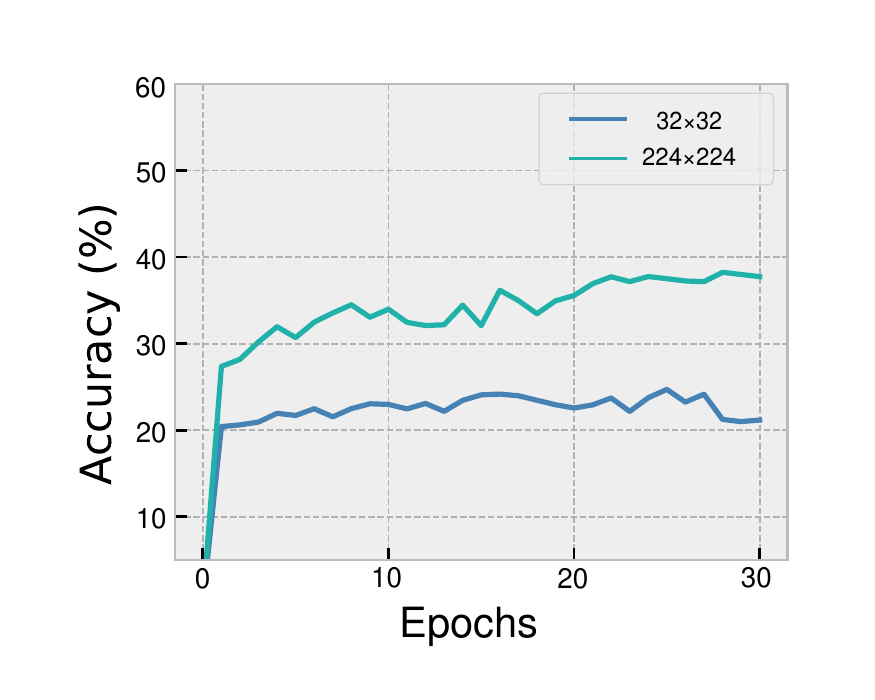}%
        \caption{%
            Different input dimensions
            on an unlearnable training set.
        }\label{fig:transfer:test}
    \end{subfigure}
    \hfill
    \begin{subfigure}{0.24\linewidth}
        \includegraphics[width=\linewidth]{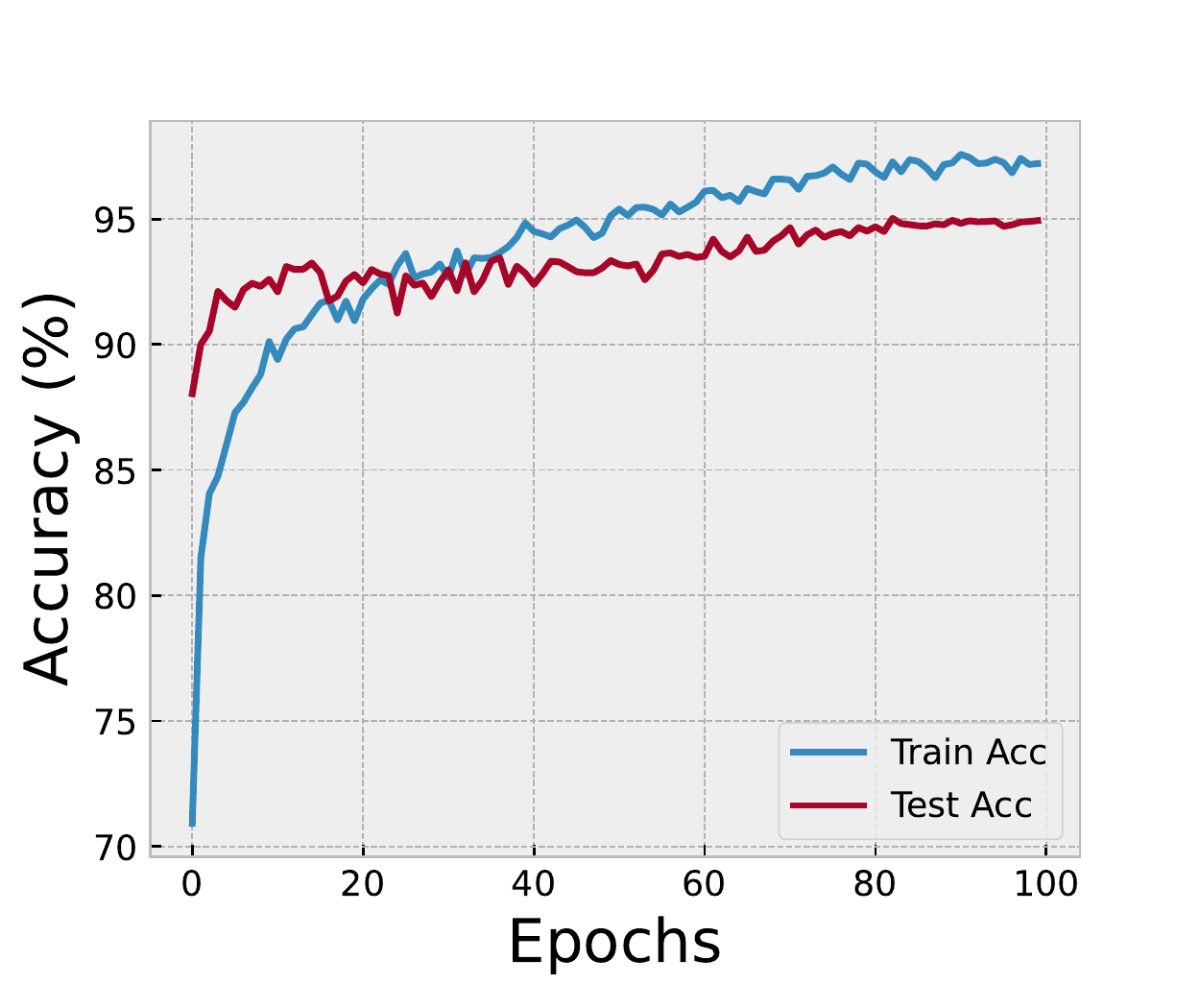}%
        \caption{%
            Train and test accuracies of transfer learning
            with \Method{}.
        }\label{fig:transfer:ueem}
    \end{subfigure}
    \caption{%
        Accuracies \wrt{} the number of training / fine-tuning epochs
        for randomly initialized / pretrained ResNet-18
        on different CIFAR-10 datasets.
        (\subref{fig:transfer:clean})
        Fine-tuning the pretrained model
        on a clean training set.
        (\subref{fig:transfer:unlearn})
        Fine-tuning the pretrained model
        with unlearnable training set
        generated with EM~\cite{huangunlearnable}.
        (\subref{fig:transfer:test})
        Comparing the test accuracies
        by training from scratch
        with either \(32 \times 32\)
        or upsampled \(224 \times 224\)
        unlearnable examples.
        (\subref{fig:transfer:ueem})
        Fine-tuning on unlearnable data
        with \Method{}.
    }\label{fig:transfer}
\end{figure}

The results of the experiments
are shown in~\Cref{fig:transfer}.
\Cref{fig:transfer:unlearn}
shows that fine-tuning
with unlearnable examples
can improve the clean test accuracy
from \(22\%\) to \(66\%\).
Additionally in~\Cref{fig:transfer:test},
we find that simply upsampling
the unlearnable samples
to use more compute and have larger feature maps
does not significantly weaken
the unlearning attack
(test accuracy increased to \(34\%\)).
Most importantly,
\Method{} successfully
eliminates the negative impact
of unlearning poisons,
which enables the model
to utilize pretrained knowledge effectively.
This enables the fine-tuned model
to achieve a test accuracy
of approximately \( 95\% \)
as shown in~\Cref{fig:transfer:ueem}.

    \section{Discussion}\label{limit}


In this section, we investigate
limitations of \Method{}
and the potential use cases of the provided options.
We compared \MethodLite{} training
with normal training
and found that \MethodLite{}
requires more training rounds
to converge due to data augmentation policies.
For instance,
CIFAR-10 typically converges
with 60 epochs of normal training,
whereas it requires more than
150 epochs of training to reach
full convergence with \MethodLite{}.
With respect to \MethodMax{}
and \Method{},
the error-maximizing augmentations
consumes the same amount
of floating-point operations (FLOPs)
as adversarial training,
if \( K \) is the same as the adversarial attack steps.
However,
we note that
our approach is easily parallelizable,
but adversarial training approaches are not,
with compute dependences between adversarial iterations.
One can choose between \MethodMax{},
\MethodLite{} and \Method{} depending
on their requirements.
\MethodMax{}
can be preferred if seeking higher clean accuracy,
whereas \Method{} or \MethodLite{}
can be a practical option
for faster training
with nearly the same accuracies.

Despite acknowledging
that a malicious party
may exploit the approach suggested in this paper,
we believe that the ethical approach
for the open-source deep learning community
is not to withhold information
but rather to increase awareness of these risks.

    \section{Conclusion}
Using the intuition
of disrupting the unlearning perturbation
with perturbations beyond the \( \ell_p \) budgets,
we propose a simple yet effective
defense method called \Method{},
which can mitigate unlearning poisons
and restore clean accuracies.
\Method{} achieves robust defenses
on unlearning poisons
with simple data augmentations
and adversarial augmentation policies.
Similar to adversarial training,
it employs error-maximizing augmentation
to further eliminate the impact
of unlearning poisons.
Our comprehensive experiments
on five state-of-the-art unlearnable example attacks
demonstrate that \Method{}
outperforms existing countermeasures
such as adversarial training~\cite{%
    huangunlearnable,furobust,tao2021better}.
We also evaluate adaptive poisons
and transfer learning on \Method{}.
Our results suggest
that existing unlearning perturbations
are tragically inadequate
in making data unlearnable.
By understanding the weaknesses of existing attacks,
we can anticipate how malicious actors
may attempt to exploit them,
and prepare stronger safeguards against such threats.
We hope \Method{}
can help facilitate the advancement
of research in these attacks and defenses.
Our code is open source
and available to the deep learning community
for scrutiny\footnote{\repourl}.

    \appendix
    \section{Experimental Setup}\label{app:setup}

\subsection{Datasets}\label{app:setup:datasets}

\textbf{CIFAR-10}
consists of 60,000 \( 32\times32 \)
resolution images,
of which 50,000 images are the training set
and 10,000 are the test set.
This dataset contains 10 classes,
each with 6000 images.

\textbf{CIFAR-100}
is similar to CIFAR-10.
It has 100 classes,
Each class has 600 images of size \( 32\times32 \),
of which 500 are used as
the training set and 100 as the test set.

\textbf{SVHN},
derived from Google Street View door numbers,
is a dataset of cropped images containing sets of
Arabic numerals `0-9'.
The dataset consists of 73,257 digit images
in the training set and 26,032 digit images in the test set.

\textbf{ImageNet-subset}
refers to a dataset constructed
with the first 100 classes from ImageNet
resized to \( 32 \times 32 \),
following the setup in~\cite{huangunlearnable}
for fair comparisons.
The training set comprises approximately 120,000 images,
and the test set contains 5,000 images.

\Cref{tab:dataset}
shows the detail specifications of these datasets.
\begin{table}[h]
\centering
\caption{%
    Overview of the specifications of datasets used in this paper.
}\label{tab:dataset}
\begin{tabular}{l|cccc}
\toprule
Dataset
    & Input size & Train-set & Test-set & Classes \\
\midrule
CIFAR-10
    & \( 32 \times 32 \times 3 \) & 50,000 & 10,000 & 10 \\
CIFAR-100
    & \( 32 \times 32 \times 3 \) & 50,000 & 10,000 & 100 \\
SVHN
    & \( 32 \times 32 \times 3 \) & 73,257  & 26,032  & 10 \\
ImageNet-subset
    & \( 32 \times 32 \times 3 \) & 127,091 & 5,000 & 100 \\
\bottomrule
\end{tabular}
\end{table}

\subsection{Models and Hyperparameters}

We evaluate \Method{}
using a standard ResNet-18~\cite{he2016deep} architecture by default,
and extend experiments
to standard ResNet-50, DenseNet-121,
and MobileNet-v2 in~\Cref{tab:arch}.
In all the experiments,
we used a stochastic gradient descent (SGD) optimizer
with a momentum of 0.9.
\Cref{%
    tab:hyperparameters:u,%
    tab:hyperparameters:uem}
provide the default hyperparameters
used to evaluate \Method{} and \emph{}
on unlearnable examples.
\begin{table}[h]
\centering\caption{%
   Default hyperparameters for \MethodLite{}.
}\label{tab:hyperparameters:u}
\begin{tabular}{l|cccc}
\toprule
Hyperparameters & CIFAR-10 & CIFAR-100 &SVHN & ImageNet-subset \\
\midrule
Learning rate \( \alpha \)
    & 0.01 & 0.01 & 0.01 & 0.01\\
Weight decay
    & 5e-4 & 5e-4 & 5e-4 & 5e-4\\
Epochs \( E \)
    & 200 & 300 & 150 & 300\\
Batch size \( B \)
    & 128 & 128 & 128 & 128\\
\bottomrule
\end{tabular}
\end{table}
\begin{table}[h]
\centering
\caption{%
   Default hyperparameters for \Method{}.
   \MethodMax{} uses the same hyperparameters
   except \( W \) equals \( E \).
}\label{tab:hyperparameters:uem}
\begin{tabular}{l|cccc}
\toprule
Hyperparameters & CIFAR-10 & CIFAR-100 & SVHN & ImageNet-subset \\
\midrule
Learning rate \( \alpha \)
    & 0.01 & 0.01 & 0.01 & 0.01 \\
Weight decay
    & 5e-4 & 5e-4 & 5e-4 & 5e-4 \\
Epochs \( E \)
    & 200 & 300 & 150 & 300 \\
Batch size \( B \)
    & 128 & 128 & 128 & 128 \\
Number of repeated sampling \( K \)
    & 5  & 5  & 5  & 5 \\
Number of error-maximizing augmentation epochs \( W \)
    & 50  & 30  & 30  & 30 \\
\bottomrule
\end{tabular}
\end{table}

\subsection{Standard Augmentation}

For CIFAR-10, SVHN, and CIFAR-100 baselines
to compare against,
we perform data augmentation
via random flipping,
and random cropping to \( 32 \times 32 \) images
on each image.
For the ImageNet-subset,
we perform data augmentation with a 0.875 center cropping,
followed with a resize to \( 32 \times 32 \),
and random flipping for each image.

\subsection{Unlearning Perturbution Budgets}\label{app:perturbation}

The attacks,
EM~\cite{huangunlearnable}, REM~\cite{furobust},
and HYPO~\cite{tao2021better},
all have a permitted perturbation bound
of \(\ell_\infty = 8/255\) for each image.
Additionally,
the LSP~\cite{yu2022availability}
and AR~\cite{sandoval2022autoregressive} attacks
permit \(\ell_2 = 1.0\).

\subsection{Adversarial Training}

For comparison,
the baseline defenses against the five methods
(EM, REM, HYPO, LSP, and AR)
on CIFAR-10
employ PGD-7 adversarial training~\cite{madry2017towards},
following the evaluation of~\cite{furobust}.
The adversarial training perturbation bounds used
were \( \ell_\infty = 4/255 \) for EM, REM and HYPO,
and \( \ell_2 = 0.5 \) for LSP and AR as baseline defenses.

\subsection{Hyperparameters for \Method{}}
\begin{figure}[ht]
    \centering
    \newcommand{\augsubfig}[2]{%
        \begin{subfigure}{0.7\linewidth}
            \centering
            \includegraphics[width=\linewidth]{#2.pdf}
            \caption{Visualization of augmented images of {#1}.}
        \end{subfigure}
    }
    \augsubfig{CIFAR-10}{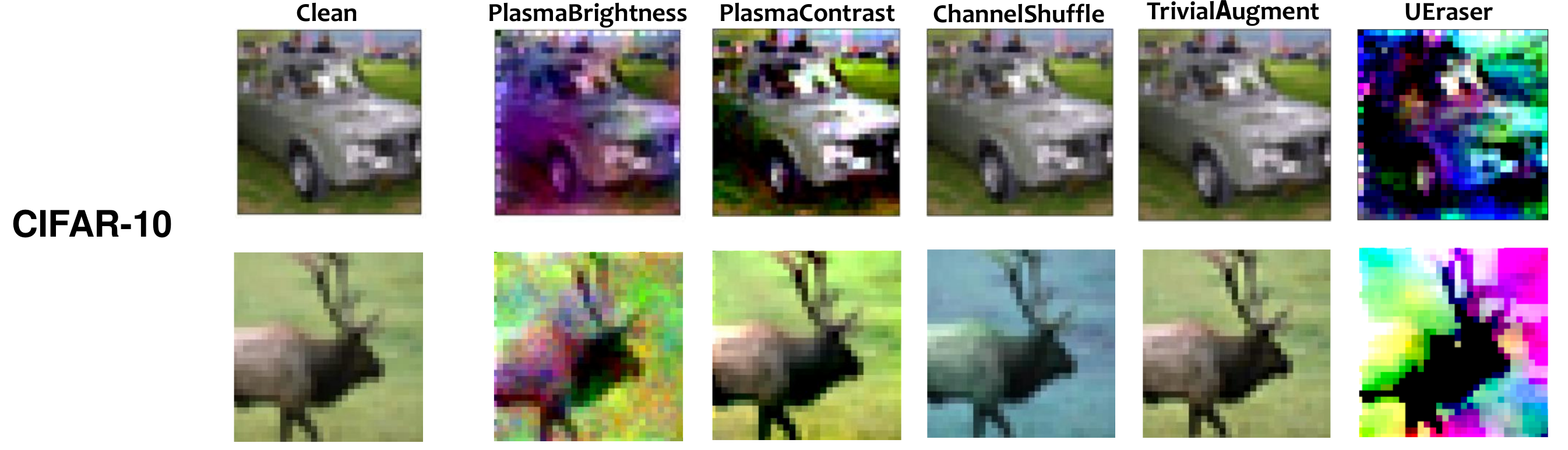} \\
    \augsubfig{CIFAR-100}{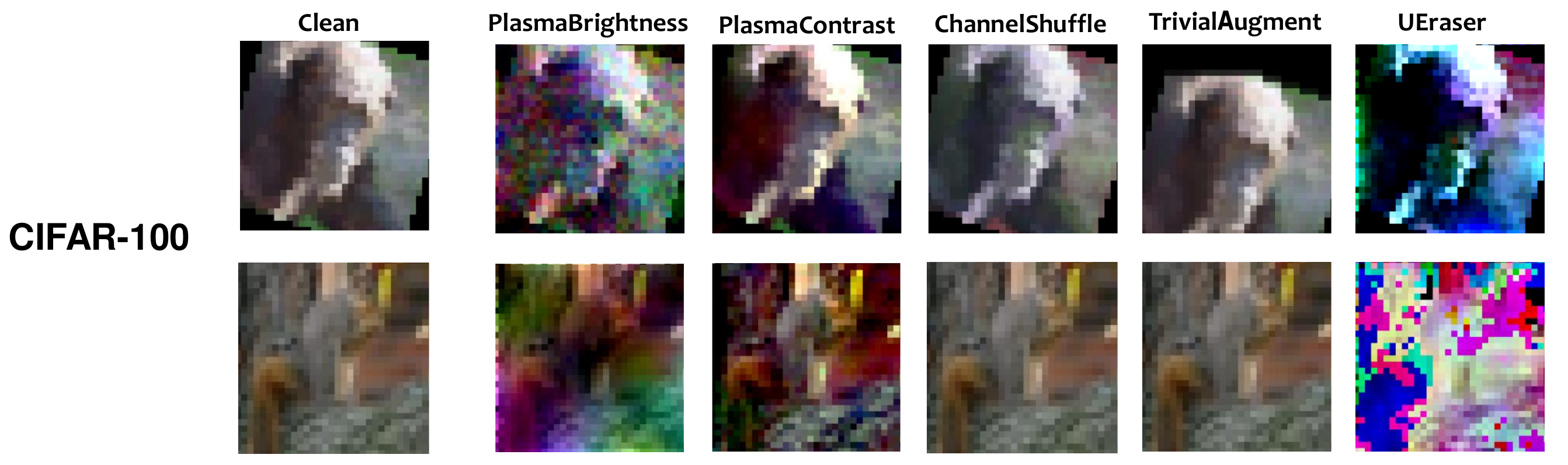} \\
    \augsubfig{SVHN}{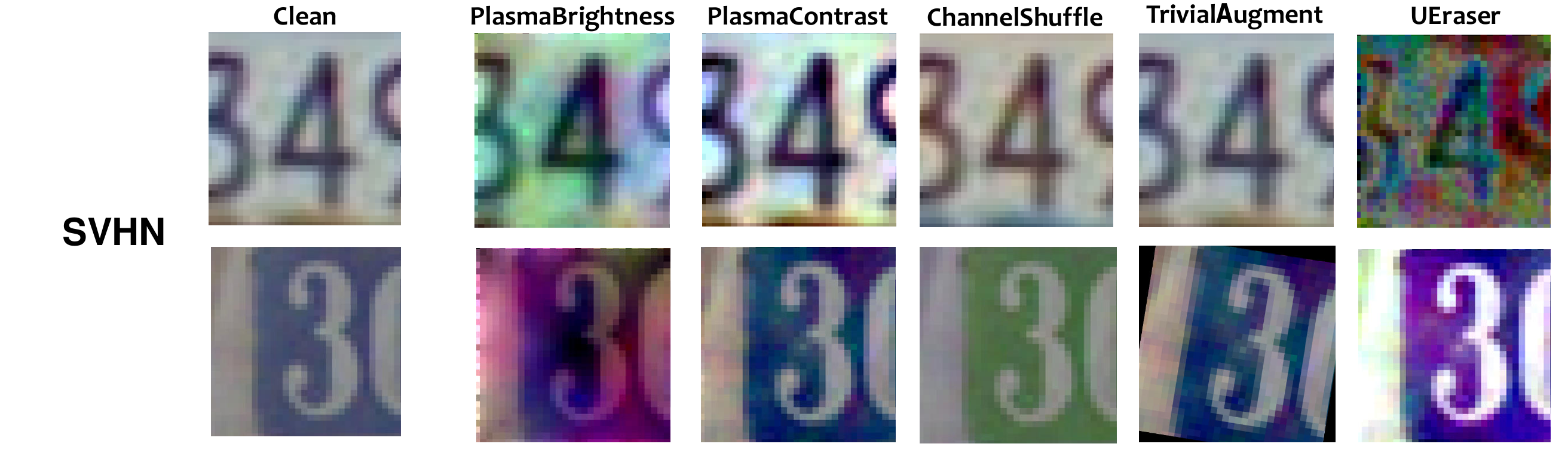} \\
    \augsubfig{ImageNet}{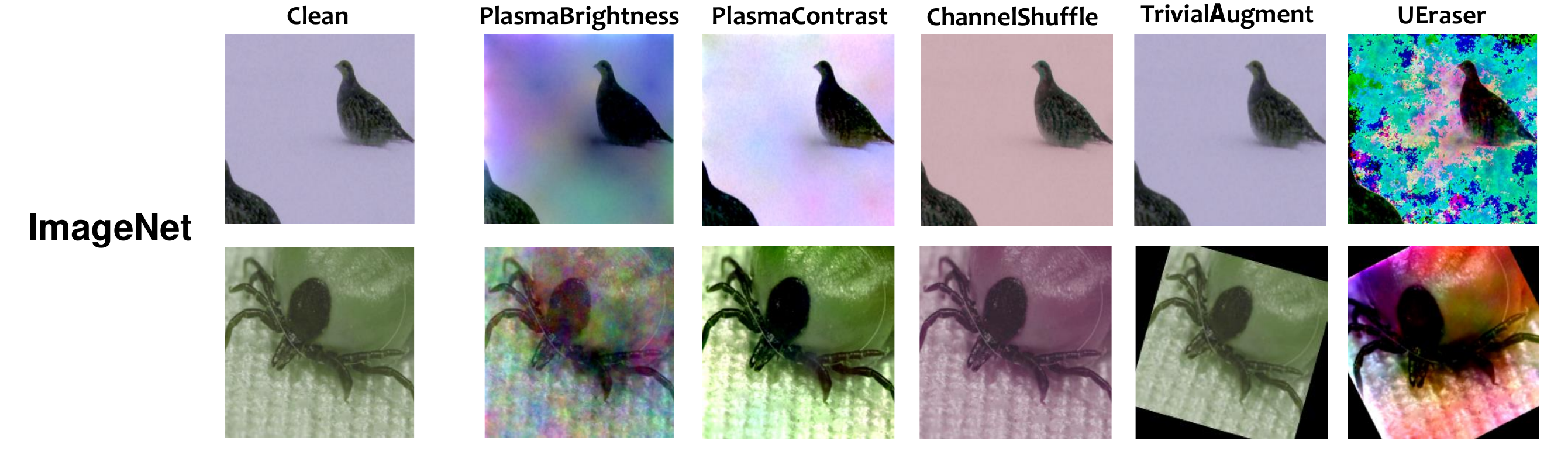} \\
    \caption{%
        The visualization of various augmentations
        on different datasets.
    }\label{fig:augs}
\end{figure}

\Method{} comprises three composite augmentations
(PlasmaTransform, ChannelShuffle, TrivialAugment).
We implement augmentations using Kornia%
\footnote{Documentation: \url{https://kornia.readthedocs.io/en/latest/augmentation.module.html}.},
specifically, with {\texttt{kornia.augmentation.RandomPlasmaBrightness}},
\texttt{kornia.augmentation.RandomChannelShuffle},
and \texttt{kornia.augmentation.auto.TrivialAugment}.
Here,
TrivialAugment uses the default hyperparameters,
\Cref{tab:hyperparameters:ua}
shows the hyperparameter settings
for the remaining augmentations.
Finally, \Cref{fig:augs}
provides the visualization of the augmentation effects
on different dataset examples.
\begin{table}[h]
\centering\caption{%
    Default hyperparameters
    for \Method{} augmentations.
}\label{tab:hyperparameters:ua}
\small
\begin{tabular}{l|ccc}
\toprule
Augmentations & PlasmaBrightness & PlasmaContrast  & ChannelShuffle \\
\midrule
Probability of use \( p \)
    & 0.5 & 0.5 & 0.5 \\
Roughness
    & (0.1, 0.7) & (0.1, 0.7) & - \\
Intensity
    & (0.0, 1.0) & - & - \\
Same on batch
    & False & False & False \\

\bottomrule
\end{tabular}
\end{table}

\section{Sensitivity Analysis}\label{app:sensitivity}

\begin{figure}[ht]
    \centering
    \newcommand{\senssubfig}[2]{%
        \begin{subfigure}{0.31\linewidth}
            \includegraphics[width=\linewidth]{#2.pdf}
            \caption{{#1}.}
        \end{subfigure}
    }
    \senssubfig{\Method{} with \( K = 1\) (\MethodLite{})}{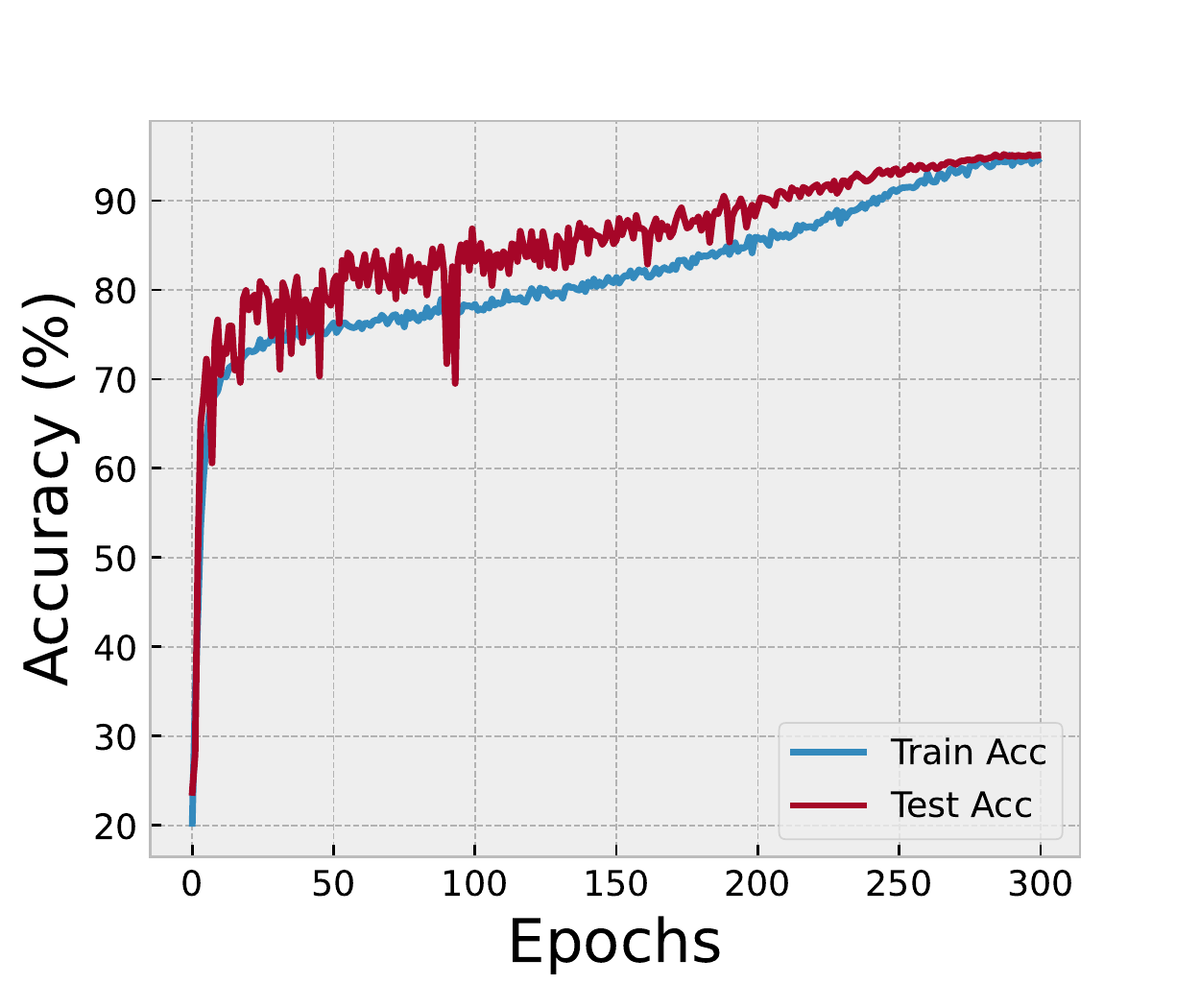}
    \senssubfig{\Method{} with \( K = 5 \)}{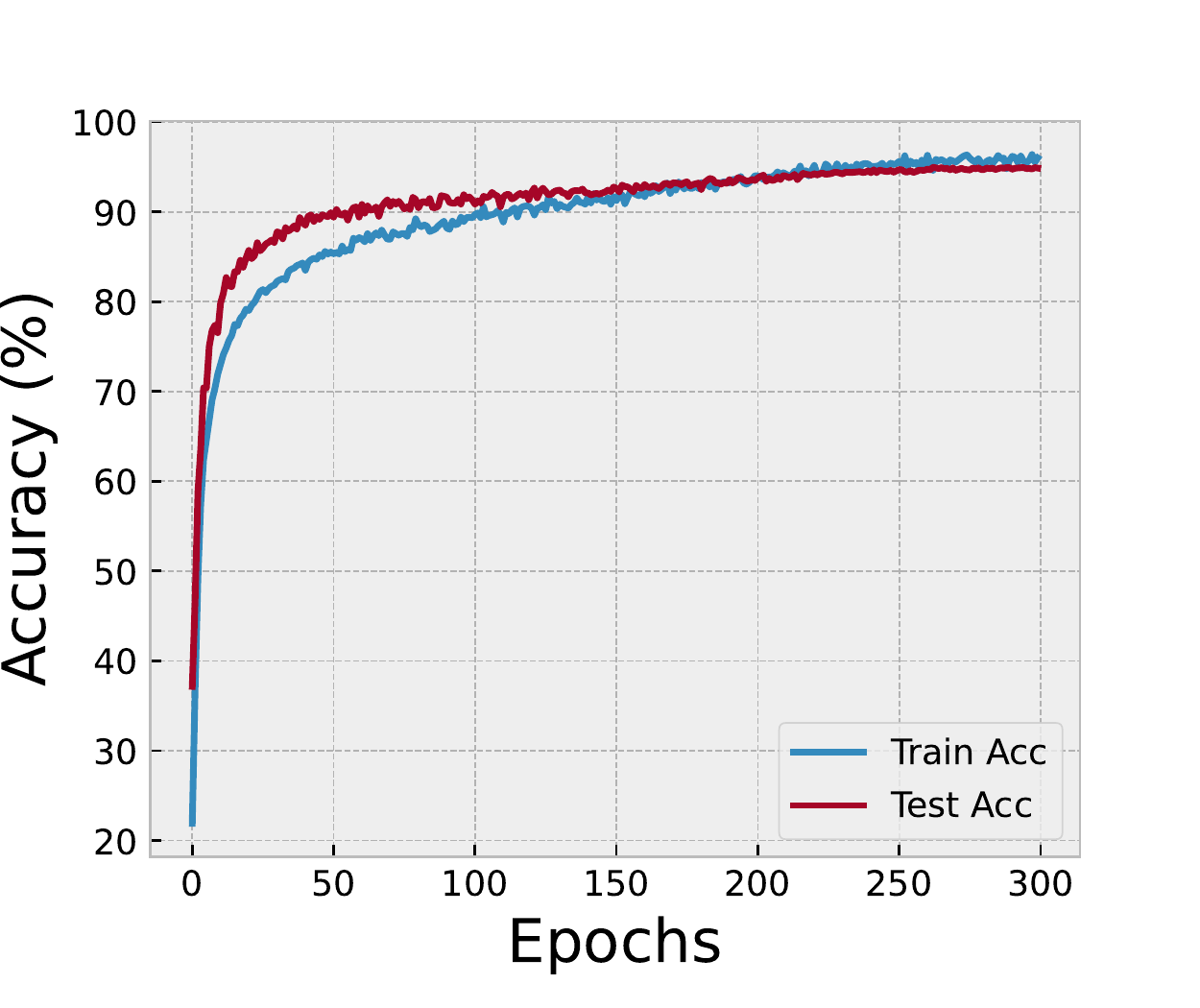}
    \senssubfig{\Method{} with \( K = 10 \)}{UEk=5}
    \\
    \senssubfig{\Method{} with \( K = 3 \)}{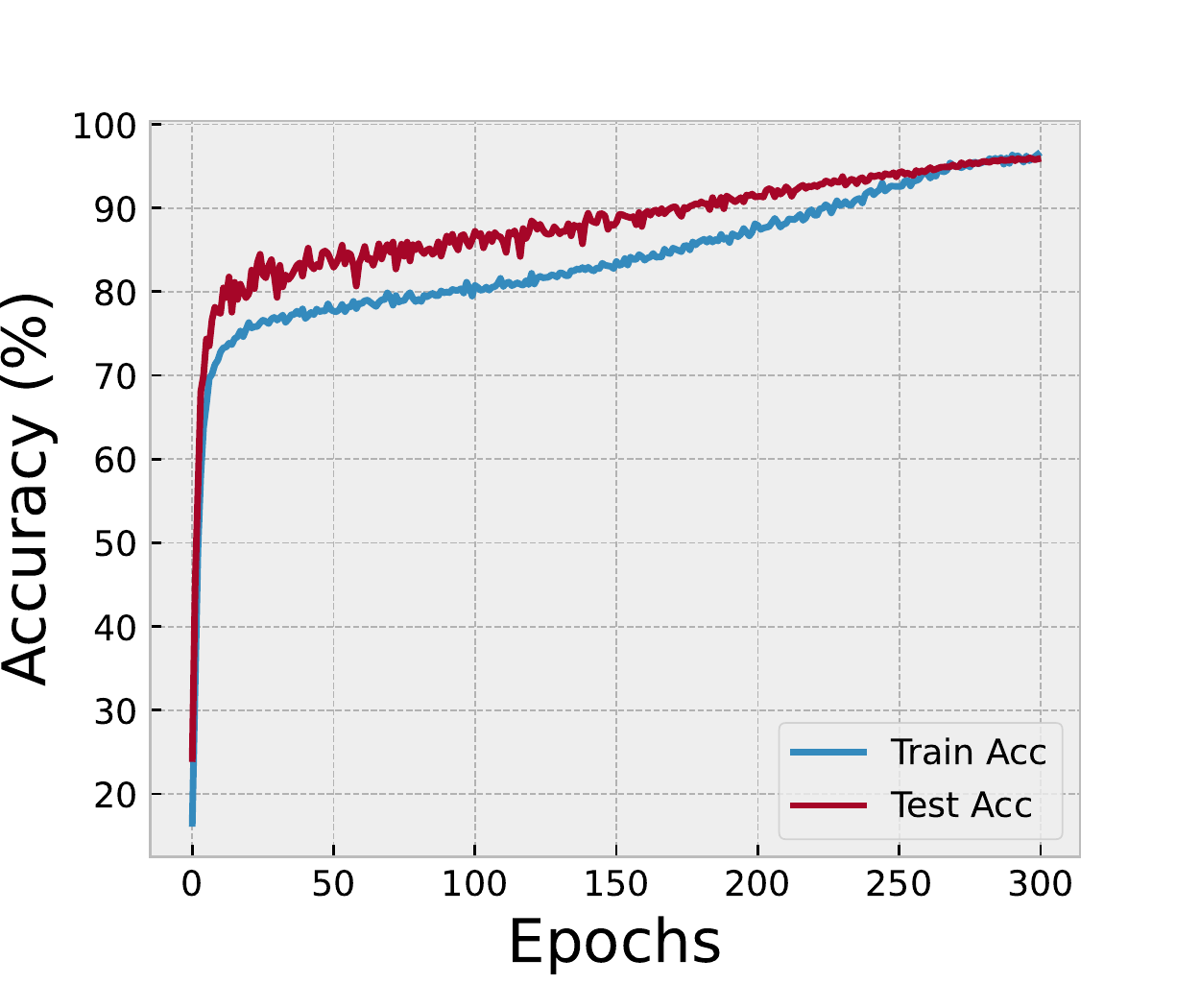}
    \senssubfig{\Method{} with \( K = 5 \)}{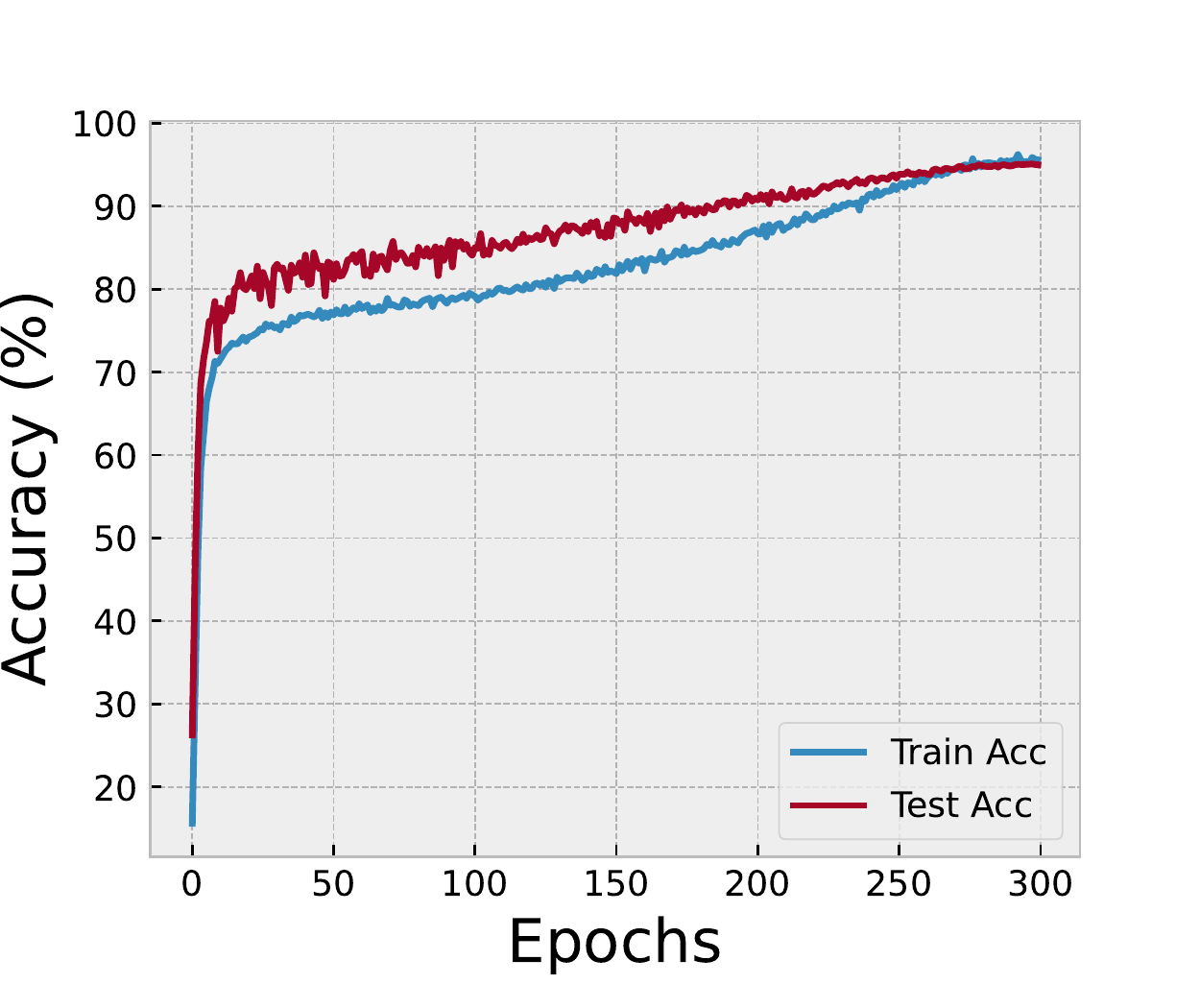}
    \senssubfig{\Method{} with \( K = 10 \)}{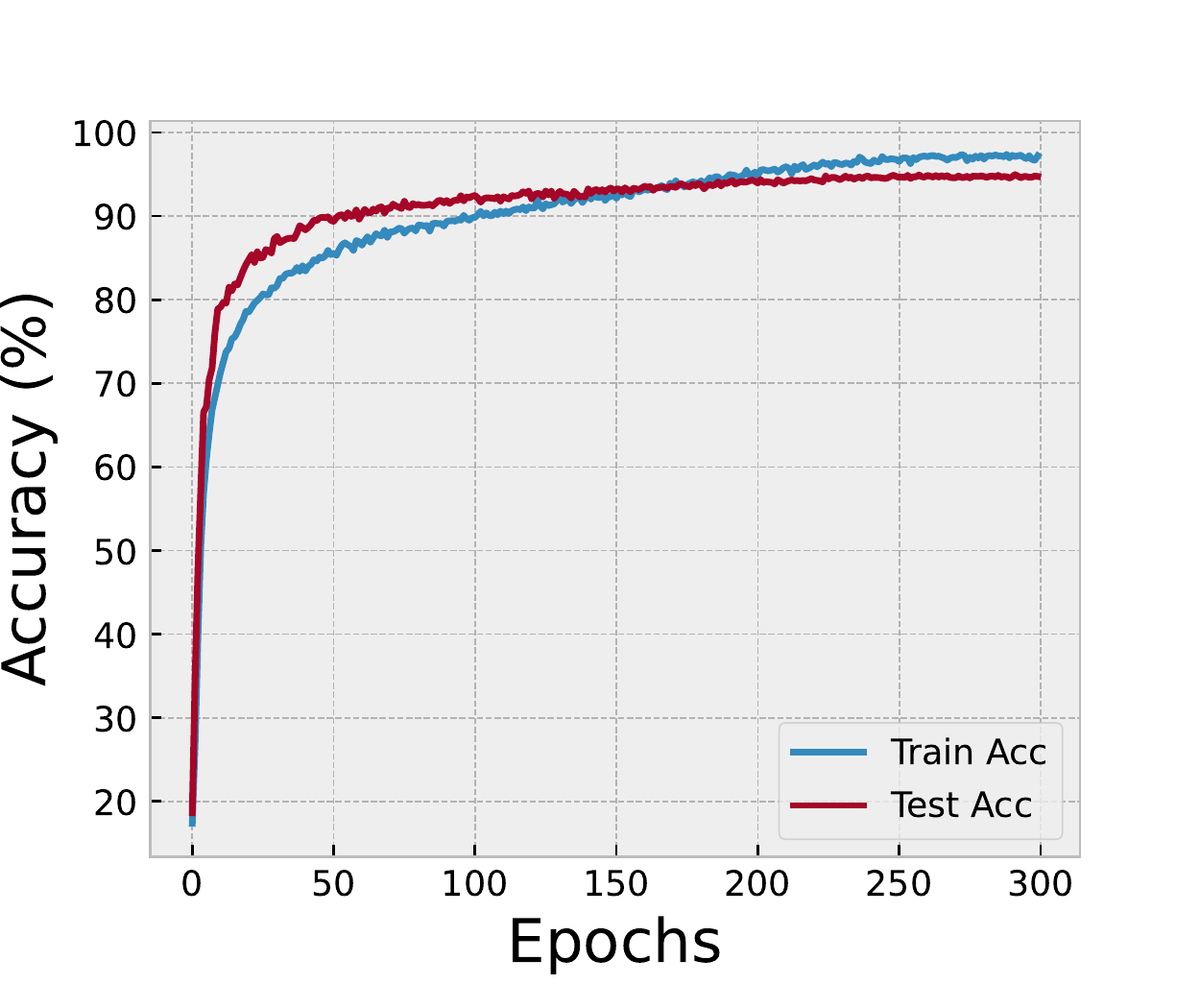}
    \caption{%
        Train and test accuracy curves
        \wrt{} the number of training epochs.
        The subfigures correspond
        to \Method{} under \( K \in \braces{1, 5, 10} \)
        and \MethodMax{} under \( K \in \braces{3, 5, 10} \).
        Recall that \( K \) is the number
        of repeated augmentation sampling.
    }\label{fig:additional_sensitivity}
\end{figure}

Here, we provide a sensitivity analysis
of the number of repeated sampling \(K\)
using in adversarial augmentation.
We also explore the effect
of increasing the number of training epochs
under \Method{}.
Taking the example of unlearnable CIFAR-10
produced with EM~\cite{huangunlearnable},
\Cref{tab:sensitivity}
shows the results of \Method{}
with various combinations
of different numbers of repeated augmentation sampling \( K \)
and the total number of training epochs \( E \).
Higher \( K \) values can effectively
improve the defense performance of \Method{},
with a training cost increasing proportionally with \( K \).
More training epochs
can also improve the performance of \Method{},
and even matches the test accuracy
of training with clean data.
Finally,
\Cref{fig:additional_sensitivity}
provides the train and test accuracy curves
\wrt{} the number of training epochs
for different \( \parens{E, K} \) configurations.
\begin{table}[h]
\centering
\caption{%
    Clean test accuracies (\%)
    of different numbers of repeated augmentation sampling
    \( K \) for \Method{}.
    Note that \( K = 1\) denotes \MethodLite{},
    and the number of error-maximizing epochs \( W = 50 \),
    ``---'' means \( K = 1 \) is not applicable for \MethodMax{}.
    The unlearnable training data
    is generated with EM on CIFAR-10,
    and a standard ResNet-18
    trained on this data attains a test accuracy
    of \( 21.21\% \).
}\label{tab:sensitivity}
\begin{tabular}{c|cc|cc}
\toprule
    \multirow{2}{*}{\( K \)}
    & \multicolumn{2}{c}{\Method{}} & \multicolumn{2}{c}{\MethodMax{}}\\
    & \( E = 200 \) & \( E = 300 \)
    & \( E = 200 \) & \( E = 300 \) \\
    \midrule
    1  & 90.78 & 94.19 &   --- &   --- \\
    2  & 92.76 & 94.79 & 91.32 & 95.25 \\
    3  & 92.91 & 95.01 & 91.40 & 95.85 \\
    4  & 93.02 & 95.14 & 91.26 & 95.59 \\
    5  & 93.38 & 94.83 & 92.12 & 95.24 \\
    6  & 93.24 & 94.78 & 93.17 & 95.36 \\
    7  & 93.16 & 94.50 & 93.58 & 94.79 \\
    8  & 93.23 & 94.87 & 93.95 & 94.84 \\
    9  & 92.86 & 94.58 & 93.92 & 94.66 \\
    10 & 93.03 & 94.83 & 94.06 & 94.57 \\
\bottomrule
\end{tabular}
\end{table}



\section{Attack and Defense Baselines}\label{app:baselines}

We use five baseline attacks
and two exisiting SOTA defenses
for evaluation and comparisons
in our experiments (\Cref{tab:source_code}).
Each attack method is implemented
from their respective official source code
for a fair comparison.
We adopt experimental setup
identical to the original publications,
and use perturbation budgets
described in~\Cref{app:perturbation}.
For defenses,
we compare \Method{} variants
against the current SOTA techniques,
image shortcut squeezing~\cite{liu2023image}
and adversarial training~\cite{madry2017towards}.
The compared defenses (ISS and adversarial training)
respectively follow the original source code
and PGD-7 adversarial training~\cite{madry2017towards}.
\begin{table}[t]
    \centering
    \caption{%
        Attack and defense methods
        and respective links to open source repositories.
    }\label{tab:source_code}
    \adjustbox{max width=\linewidth}{%
    \begin{tabular}{ll}
        \toprule
        Name & Open Source Repository \\
        \midrule
        \multicolumn{2}{c}{Attacks} \\
        \midrule
        Error-minimizing attack (EM)~\cite{huangunlearnable}
        & \url{https://github.com/HanxunH/Unlearnable-Examples/} \\
        Robust error-minimizing attack (REM)~\cite{furobust}
        & \url{https://github.com/fshp971/robust-unlearnable-examples} \\
        Hypocritical perturbations (HYPO)~\cite{tao2021better}
        & \url{https://github.com/TLMichael/Delusive-Adversary} \\
        Linear-separable synthetic perturbations (LSP)~\cite{yu2022availability}
        & \url{https://github.com/dayu11/Availability-Attacks-Create-Shortcuts} \\
        Autoregressive poisoning~\cite{sandoval2022autoregressive}
        & \url{https://github.com/psandovalsegura/autoregressive-poisoning} \\
        \midrule
        \multicolumn{2}{c}{Defenses} \\
        \midrule
        Image shortcut squeezing~\cite{liu2023image}
        & \url{https://github.com/liuzrcc/ImageShortcutSqueezing} \\
        Adversarial Training~\cite{madry2017towards}
        & Example implementation from \url{https://github.com/fshp971/robust-unlearnable-examples} \\
        \bottomrule
    \end{tabular}}
\end{table}

    \FloatBarrier
    {\small
    \bibliographystyle{ieee_fullname}
    \bibliography{references}
    }
\end{document}